\documentclass[sigconf,nonacm]{acmart}


\renewcommand{\ClassWarning}[2]{}

\usepackage{booktabs} 
\usepackage{subcaption}

\usepackage{array}
\usepackage{wrapfig}
\usepackage{multirow}

\usepackage[inline]{enumitem}

\usepackage{array}  

\newcommand{\revision}[1]{#1} 

\usepackage[ruled]{algorithm2e} 

\usepackage{fancyhdr}
\setcopyright{none}
\acmDOI{}
\acmISBN{}

\usepackage[ruled]{algorithm2e} 

\SetAlFnt{\small}
\SetAlCapFnt{\small}
\SetAlCapNameFnt{\small}
\SetAlCapHSkip{0pt}

\acmJournal{TOG}

\begin{document}

\title{MatCLIP: Light- and Shape-Insensitive Assignment of PBR Material Models} 

\author{Michael Birsak}
\affiliation{%
 \institution{KAUST}
 \country{KSA}
}

\author{John Femiani}
\affiliation{%
 \institution{Miami University}
 \country{USA}
}

\author{Biao Zhang}
\affiliation{%
 \institution{KAUST}
 \country{KSA}
}

\author{Peter Wonka}
\affiliation{%
 \institution{KAUST}
 \country{KSA}
}

\begin{teaserfigure}
  \centering
  \includegraphics[width=0.95\textwidth,]{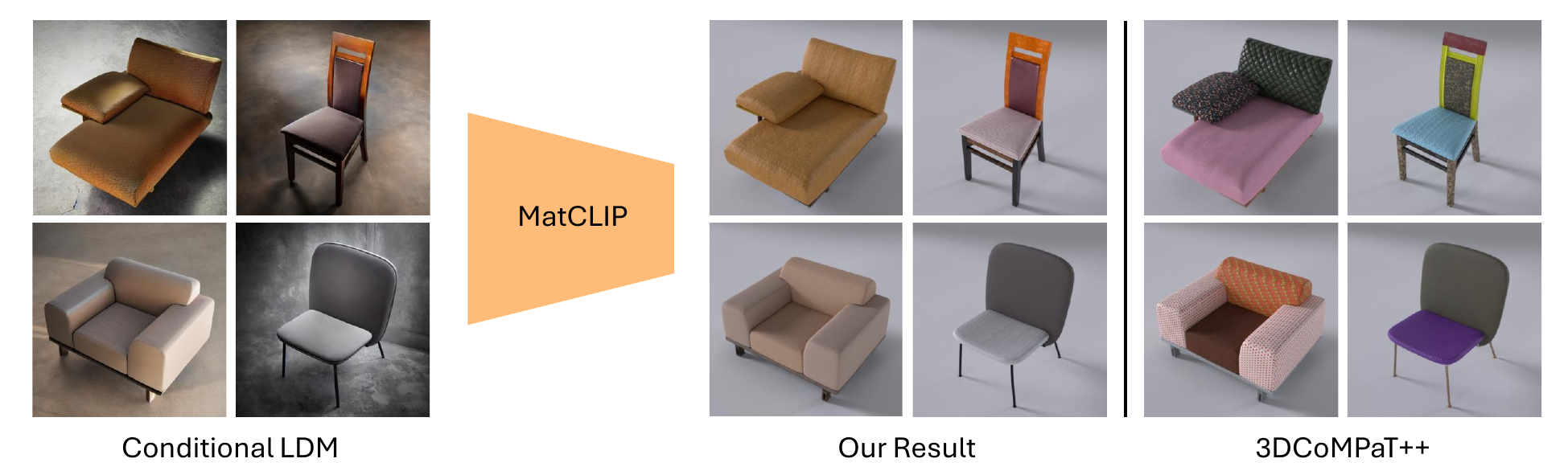}
  \caption{(Left) Latent Diffusion Models (LDMs) are a powerful tool to obtain good material assignments, but the output is RGB only. (Middle) Our MatCLIP model robustly matches a PBR material from a material database to each part of the shape. (Right) By comparison, procedurally assigning PBR materials from meaningful substance categories to 3D shapes yields unconvincing results. }
  \label{fig:teaser}
\end{teaserfigure}

\begin{abstract}



Assigning realistic materials to 3D models remains a significant challenge in computer graphics.
We propose MatCLIP, a novel method that extracts shape- and lighting-insensitive descriptors of Physically Based Rendering (PBR) materials to assign plausible textures to 3D objects based on images, such as the output of Latent Diffusion Models (LDMs) or photographs. 
Matching PBR materials to static images is challenging because the PBR representation captures the dynamic appearance of materials under varying viewing angles, shapes, and lighting conditions. 
By extending an Alpha-CLIP-based model on material renderings across diverse shapes and lighting, and encoding multiple viewing conditions for PBR materials, our approach generates descriptors that bridge the domains of PBR representations with photographs or renderings, including LDM outputs. This enables consistent material assignments without requiring explicit knowledge of material relationships between different parts of an object.
MatCLIP achieves a top-1 classification accuracy of 76.6\%, outperforming state-of-the-art methods such as PhotoShape and MatAtlas by over 15 percentage points on publicly available datasets. Our method can be used to construct material assignments for 3D shape datasets such as ShapeNet, 3DCoMPaT++, and Objaverse. All code and data will be released at \url{https://birsakm.github.io/matclip/}.

\end{abstract}

%

%
%



\maketitle

\section{Introduction}

Creating realistic 3D models involves shaping the geometry and assigning appropriate materials and textures. While many collections of high-quality 3D shapes are available, the challenge of assigning materials that enhance realism and coherence persists. Current methods for material assignments, such as automatically selecting from a material library, often result in unsatisfactory and incoherent outcomes. 
While these procedural methods ensure material assignments adhere to predefined substance categories, they lack the ability to consider interdependencies between parts or align assignments with geometric and stylistic properties (cf. Fig. \ref{fig:teaser}, right). Our method addresses these limitations by leveraging diffusion models and robust embeddings to produce coherent, contextually appropriate material assignments.
Another approach involves matching 3D shapes to stock photos with similar textures, which can result in geometric mismatches and incorrect material assignments. Additionally, one is limited to the pre-existing designs in the stock photos. Generative models allow the exploration of a broader and less constrained space of images, a feature we aim to exploit. 

Conditional Latent Diffusion Models (LDMs)~\cite{zhang2023_controlnet, rombach2021_stablediffision}, such as Stable Diffusion and Flux, have shown promise in generating plausible RGB images from a depth image of a 3D shape and a textual prompt describing the desired appearance and materials. However, these models produce a static image and not a 3D model. They cannot directly generate texture-based materials, such as Physically Based Rendering (PBR) materials often used in computer graphics, to produce high-quality models suitable for real-time rendering.

\begin{figure*} 
    \centering
    \includegraphics[width=\textwidth,trim={0 5.2cm 0 5.2cm},clip]{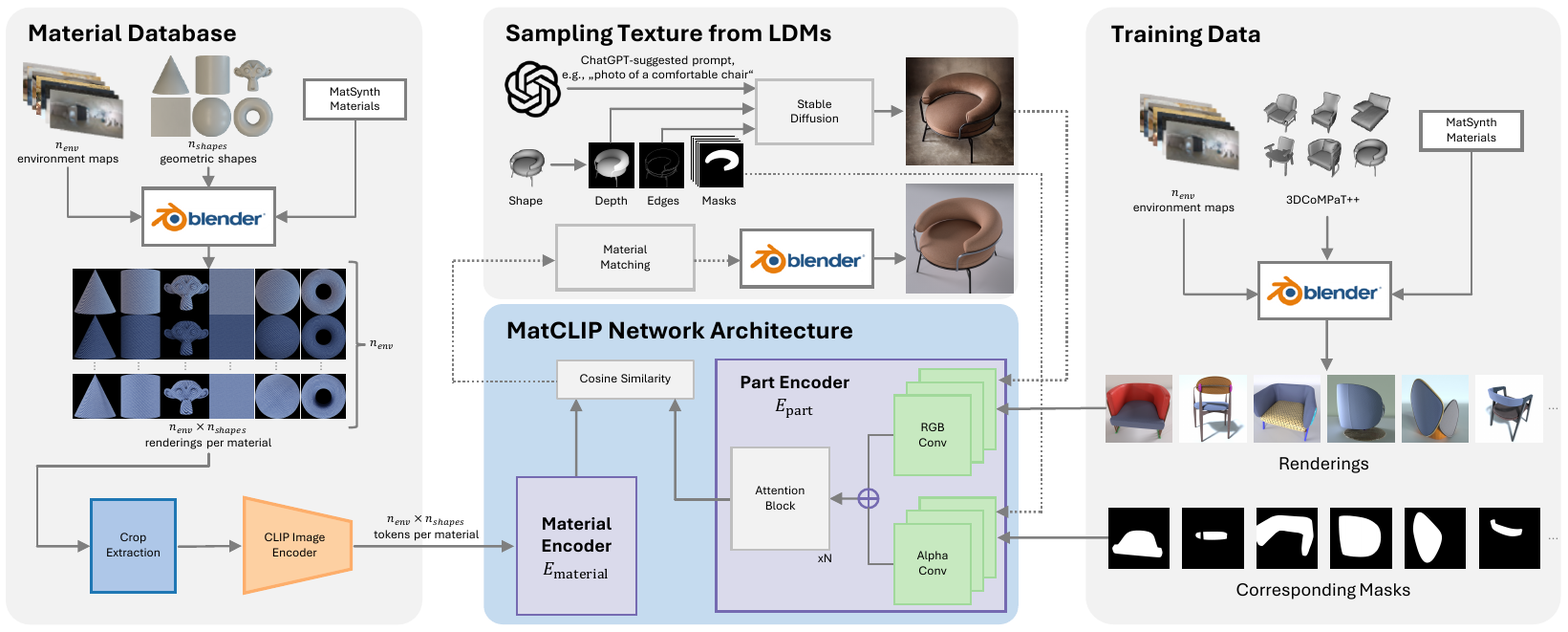}
    \caption{(left) Our method learns robust material descriptors through diverse renderings across geometric shapes and lighting conditions. (bottom center) MatCLIP aligns material embeddings with corresponding parts of 3D objects using cosine similarity and attention mechanisms. (right) We train our MatCLIP model using a large dataset of masked renderings. (top center) We leverage the output from LDMs (e.g., Stable Diffusion) to obtain meaningful material assignments to untextured shape collections.}
    \label{fig:overview}  
\end{figure*}

To address these limitations, we propose a novel material classification model by extending Alpha-CLIP. Assigning materials from an existing material database has several advantages. First, several sources for high-quality and high-resolution texture-based PBR materials are available.
For our material database, we leverage MatSynth \cite{vecchio2023matsynth}, a publicly available resource containing over 4,000 high-resolution (4K) PBR materials across diverse categories,  including maps for basecolor, diffuse, normal, height, roughness, metallic, specular, and opacity.
Second, materials from diffusion-based PBR texture generators like MatFuse~\cite{matfuse2024} are usually unable to generate textures at such high resolutions, or they struggle with generating maps with a similar level of geometric detail.

Our contributions are as follows:
\begin{enumerate}[label=(\roman*)]
    \item We introduce a novel method for learning embeddings of PBR materials that are robust to shape and lighting. These embeddings, trained on diverse renderings, enable robust and consistent material assignment across varied geometries and lighting conditions.
    \item We provide an open and reproducible framework that relies on public datasets (MatSynth and 3DCoMPaT++) and foundational diffusion models, ensuring accessibility and transparency for the community.
    \item We achieve state-of-the-art performance in material assignment, with a Top-1 accuracy of 76.69\% on MatSynth, surpassing prior methods such as PhotoShape and MatAtlas.
\end{enumerate}

\section{Related Work}\label{sec:related-work}

Creating realistic materials for 3D models is a longstanding challenge in computer graphics, addressed through diverse approaches such as exemplar-based material assignment, RGB-only texturing, SVBRDF estimation, and diffusion-based material generation. Methods based on exemplars like PhotoShape \cite{photoshape2018} classify material categories using photographs but lack robust embeddings for PBR materials, limiting generalization. Extensions like PhotoScene \cite{photoscene2022} optimize material parameters through differentiable graphs. 
\revision{The best-matching material graph is identified by nearest neighbor search using a VGG feature distance. Similarly, PSDR-Room~\cite{psdr2023} and MaPa~\cite{mapa2024} utilize CLIP in a zero-shot manner between pre-rendered thumbnails of material graphs and image crops to obtain best-matching material graphs. These methods depend on the proprietary Adobe Substances material library, limiting external reproducibility by the community. The Visual Language of Fabrics (VLF) \cite{deschaintre23_visual_fabric} fine-tunes CLIP on a curated dataset of (rendering, description) pairs exclusively for fabrics to improve the quality of text-based and image-based retrieval.} 
In contrast, our method uses contrastive learning to align embeddings across diverse lighting and geometric conditions, improving robustness and accessibility. \revision{ We want to emphasize that the idea of doing CLIP-based material retrieval has also been adopted by the methods cited above. The novelty of our approach is that we find an embedding for materials that is robust to geometry and lighting conditions but still within a CLIP space.}  

Recent texture generation techniques, such as Text2Tex \cite{text2tex2023}, TEXTure \cite{texture2023}, and MatAtlas \cite{matatlas2024}, leverage depth-conditioned diffusion models to generate RGB textures, but face challenges with consistency across parts and geometric mismatches. Similarly, SVBRDF estimation methods, including MaterialGAN \cite{materialgan2020} and SurfaceNet \cite{surfacenet2021}, infer reflectance parameters but require complex architectures and extensive training. Our method sidesteps these challenges by directly assigning materials from a curated database, avoiding the computational overhead of texture generation and reflectance estimation.

Generative approaches like ControlMat \cite{controlmat2023} and MatFuse \cite{matfuse2024} generate new textures, with ControlMat effectively addressing high-resolution outputs but lacking emphasis on geometric accuracy, while MatFuse struggles with resolution and consistency for complex geometries. SAMa \cite{fischer2024sama} and MaterialPicker \cite{ma2024materialpicker} focus on video-based diffusion models for material assignments. FlashTex \cite{flashtex2024} and PaintIt \cite{paintit2024} introduce additional complexity by requiring multi-view inputs or differentiable rendering. Unlike these methods, we emphasize reproducibility and practicality by leveraging publicly available datasets like MatSynth \cite{vecchio2023matsynth} and 3DCoMPaT++ \cite{li20223d_compat,slim_3dcompatplus_2023}, a shape dataset with over 10 million stylized 3D shapes across 41 categories.

StyleTex \cite{styletex2024} uses diffusion models with content-negative prompts and ControlNet for stylized texture generation, focusing on aesthetics over physically-based materials. TexOct \cite{liu2024texoct} synthesizes textures directly in 3D space using octree representations, addressing UV mapping and occlusion but lacking the simplicity and realism of pre-existing high-resolution PBR materials, as used in our approach. Meta 3D TextureGen \cite{meta2024texturegen} employs a two-step text-to-image and UV refinement pipeline for globally consistent textures, but it does not leverage pre-existing high-resolution PBR materials, as our method does, ensuring consistency and realism without iterative refinement. TEXGen \cite{texgen2024} incorporates hybrid 2D-3D denoising for UV texture synthesis, addressing geometric inconsistencies but retaining the computational overhead of UV mapping, which our method avoids by directly aligning materials with 3D shape parts. FlexiTex \cite{jiang2024flexitex} integrates normal maps into diffusion models for geometry-aware texture alignment but requires model-specific fine-tuning, whereas our method achieves generalization and consistency across diverse shapes using robust, pre-trained embeddings and publicly available datasets.

Our approach provides a scalable solution for assigning PBR materials, combining robust embeddings with shape- and lighting-invariant representations. By addressing limitations in material classification and ensuring coherence across diverse conditions, we position MatCLIP as a practical alternative to state-of-the-art methods for real-world applications in computer graphics.

\section{Method}\label{sec:Method}

We begin by introducing the database of PBR materials used 
in this study. Unlike prior work, we rely on publicly available 
data to ensure repeatability and accessibility. This limits 
the variety of PBR materials but ensures the approach remains 
applicable to widely accessible resources.
Next, we describe the representation of materials. We employ two different types of representations. Each shape part 
is modeled as a single CLIP-like vector, while each material 
is represented as a set of vectors capturing its appearance variations under different lighting and surface geometry.
We then discuss the training process. It uses two encoders: 
one processes images of masked parts within multipart textured 
objects (Fig.~\ref{fig:overview}, right); the other handles 
the diverse representations of PBR materials rendered under 
varying conditions (Fig.~\ref{fig:overview}, left). 
Finally, we leverage pretrained diffusion models to guide material 
selection, utilizing their latent understanding of interdependent 
material assignments across parts and constraining the process to 
align with the specific geometry and stylistic features of the 3D model.

\subsection{Material Database}

A PBR material captures a diverse set of 
potential appearances, requiring a robust representation that 
is less sensitive to lighting and shape. To achieve this, we train 
using samples rendered under diverse conditions. 

We use $n_{\text{env}} = 7$ HDR environment maps from Poly Haven 
to represent different lighting setups for studio scenarios. 
Since our study focuses on furniture (chairs), we select 
HDR images that resemble typical indoor lighting. We further 
choose $n_{\text{shapes}} = 6$ base shapes, including cones, 
cylinders, planes, spheres, Suzanne (an ape model), and tori, to ensure that material 
representations are not overly sensitive to shape variations. This results 
in $n_{\text{env}} \times n_{\text{shapes}} = 42$ renderings per material. 
For chairs, MatSynth includes 5 applicable substance categories, such as 
fabric and leather, which span a total of 1,600 materials. 
By rendering 
each material across all 42 lighting and shape combinations, we generate 
67,200 images, capturing various lighting and geometric 
conditions while maintaining computational efficiency.

Using the MatSynth material database, we render each material 
with Blender's physically-based path tracer. Each shape uses a 
fixed UV-mapping. This choice is practical, as rendering is 
computationally expensive. Additionally, consistent UV-mappings 
ensure physical scale across materials, avoiding the need to make 
the model robust to varying UV-mappings, which would require 
significant additional complexity.

We render each shape from a frontal view using a focal length 
of $100mm$, chosen to match the focal lengths commonly used in 
professional product photography, such as IKEA catalogs. The 
distance is set to maximize the shape’s extent within the image 
margins, ensuring that material details are clearly visible. 
Figure \ref{fig:overview} (left) shows a subset of renderings 
for one material applied to various shapes under various 
environment maps.

We restrict our evaluation to chairs in the 3DCoMPaT++~\cite{slim_3dcompatplus_2023} 
dataset, but the pipeline is suitable for all furniture and other products used in daily life. Chairs are well-suited for studying material assignments due to their geometric 
diversity and the presence of distinct material groups. This focus 
aligns with prior work, such as PhotoShape~\cite{photoshape2018}, 
which also concentrates on chairs. 

\paragraph{Training Data}
To generate the dataset (cf. Fig. \ref{fig:overview} (right)), we first randomly select a shape, 
then assign random materials to its parts based on their 
annotated substances. Each shape may appear multiple times 
with different material configurations.
Images are rendered from an informative perspective, selected randomly from a range of viewpoints so that key features are likely to be visible.
We apply random rotations, 
sampling typical viewing angles with $360^\circ$ azimuth and 
$0^\circ$--$90^\circ$ elevation (no roll). Models are rendered 
under the same set of HDR environment maps used in the material 
database. For each 
image, we also render a mask for each material group defined 
in the 3DCoMPaT++ model.
This process yields a dataset of 53,000 rendered images. However, 
each image contains multiple material groups, with masks indicating 
the parts where specific materials are applied. The total dataset 
includes over 200,000 individual parts (each a material group within 
an image), which serve as training samples.

\begin{figure}[t]
  \centering
    \begin{subfigure}[b]{0.09\textwidth}
        \includegraphics[width=\textwidth]{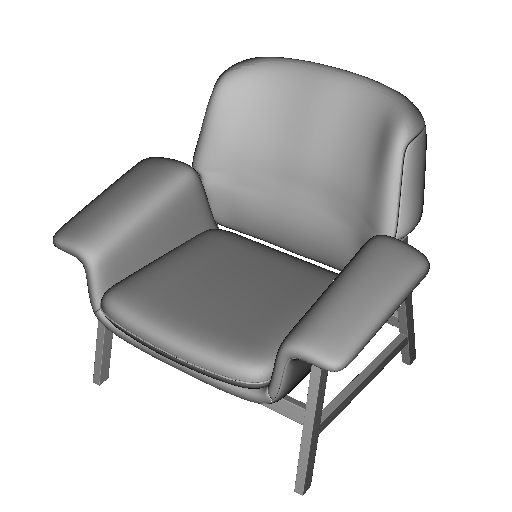}
    \end{subfigure}
    \begin{subfigure}[b]{0.09\textwidth}
        \includegraphics[width=\textwidth]{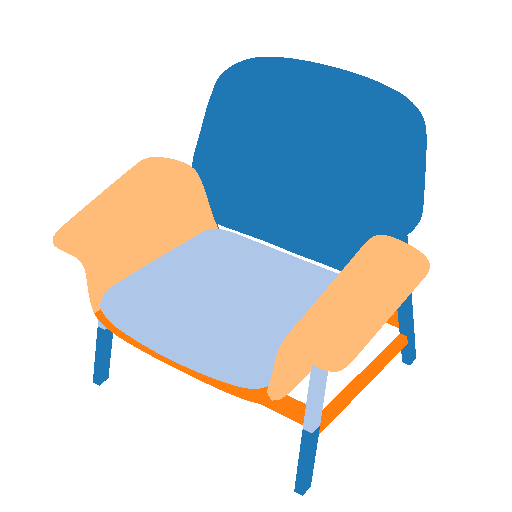}
    \end{subfigure}
    \begin{subfigure}[b]{0.09\textwidth}
        \includegraphics[width=\textwidth]{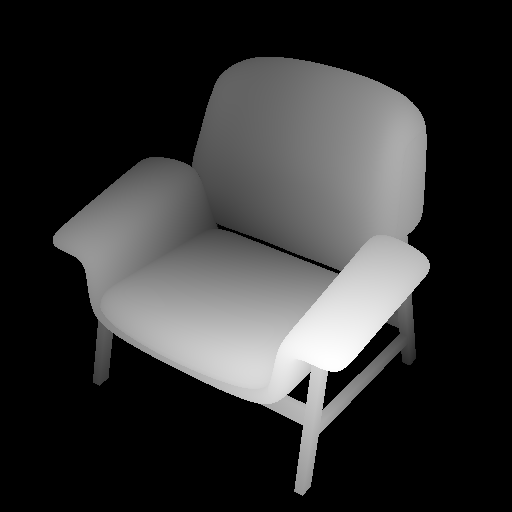}
    \end{subfigure}
    \begin{subfigure}[b]{0.09\textwidth}
        \includegraphics[width=\textwidth]{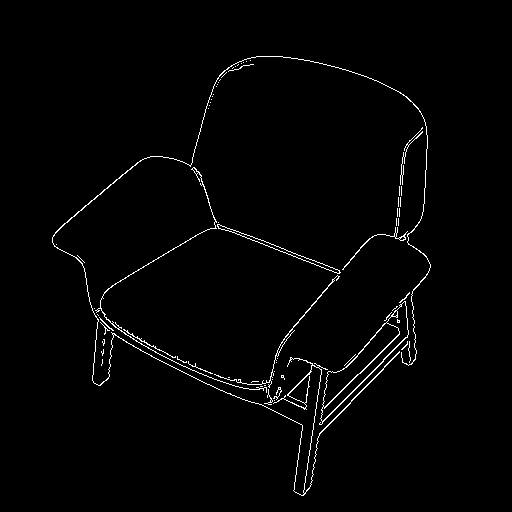}
    \end{subfigure}
    \begin{subfigure}[b]{0.09\textwidth}
        \includegraphics[width=\textwidth]{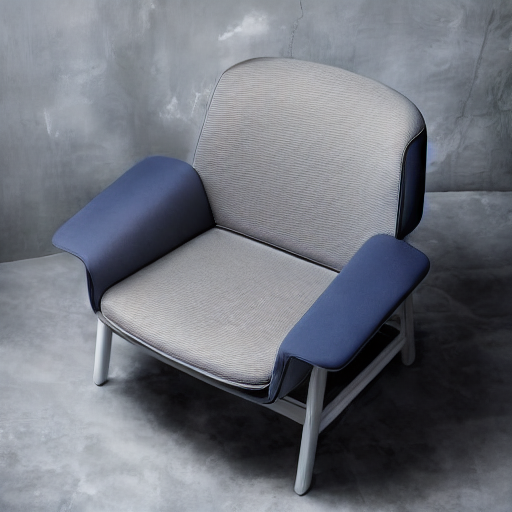}
    \end{subfigure}

    \begin{subfigure}[b]{0.09\textwidth}
        \includegraphics[width=\textwidth]{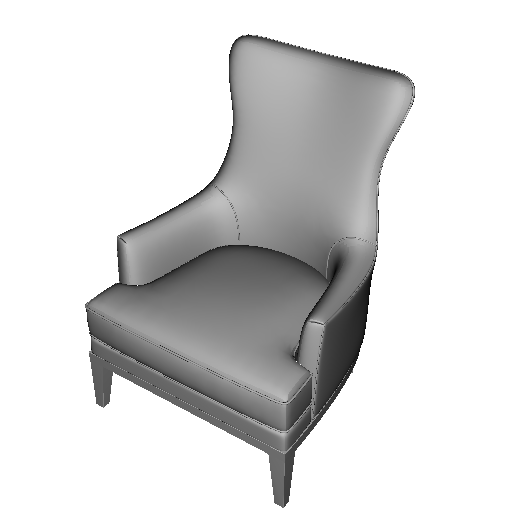}
    \end{subfigure}
    \begin{subfigure}[b]{0.09\textwidth}
        \includegraphics[width=\textwidth]{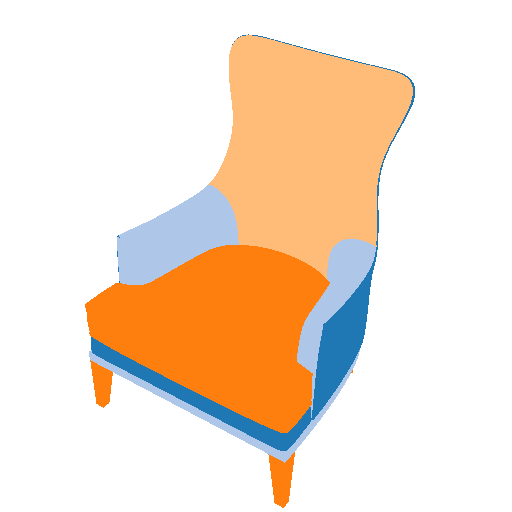}
    \end{subfigure}
    \begin{subfigure}[b]{0.09\textwidth}
        \includegraphics[width=\textwidth]{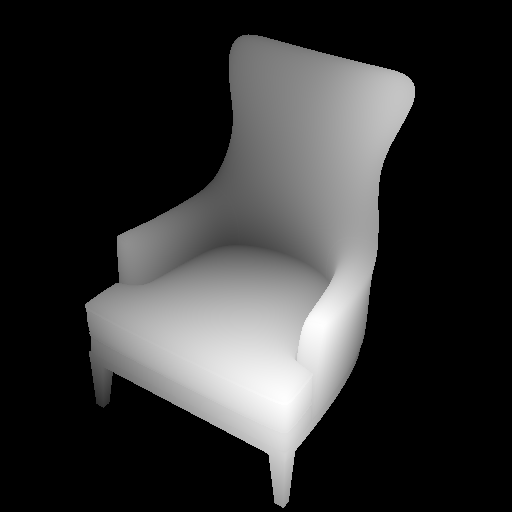}
    \end{subfigure}
    \begin{subfigure}[b]{0.09\textwidth}
        \includegraphics[width=\textwidth]{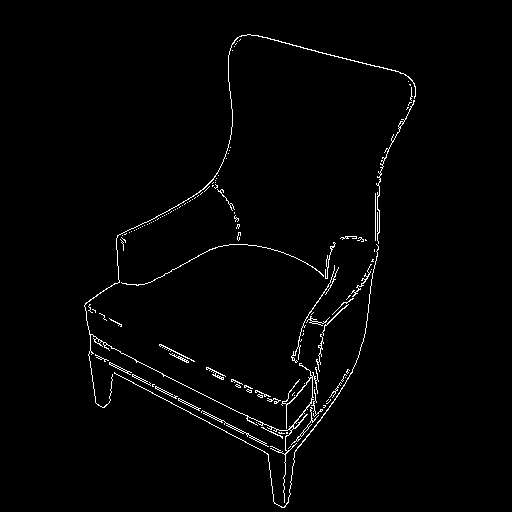}
    \end{subfigure}
    \begin{subfigure}[b]{0.09\textwidth}
        \includegraphics[width=\textwidth]{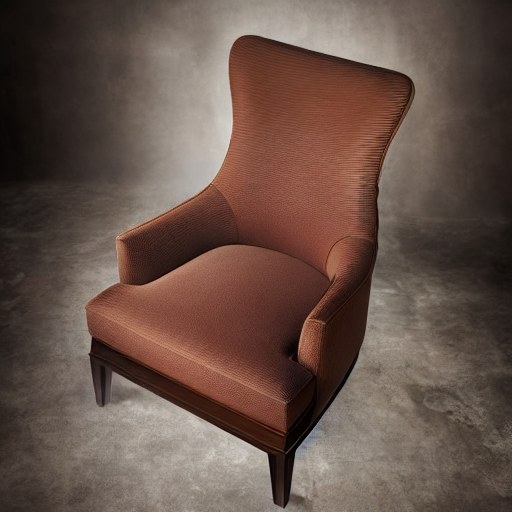}
    \end{subfigure}

    \begin{subfigure}[b]{0.09\textwidth}
        \includegraphics[width=\textwidth]{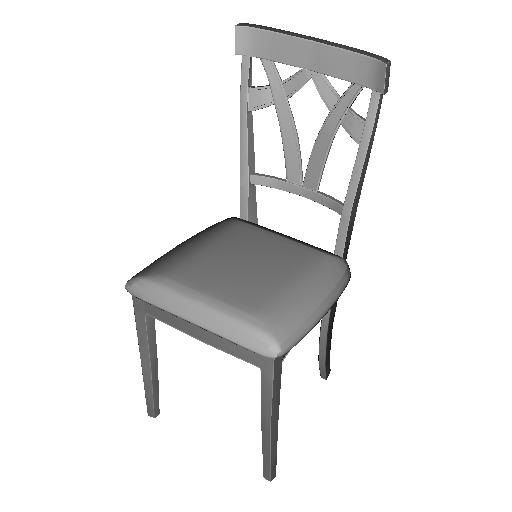}
    \end{subfigure}
    \begin{subfigure}[b]{0.09\textwidth}
        \includegraphics[width=\textwidth]{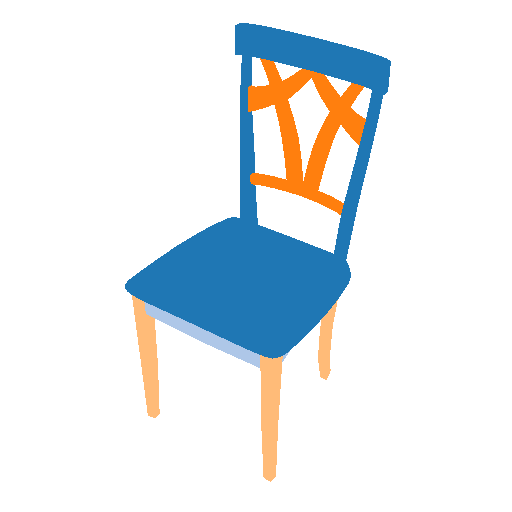}
    \end{subfigure}
    \begin{subfigure}[b]{0.09\textwidth}
        \includegraphics[width=\textwidth]{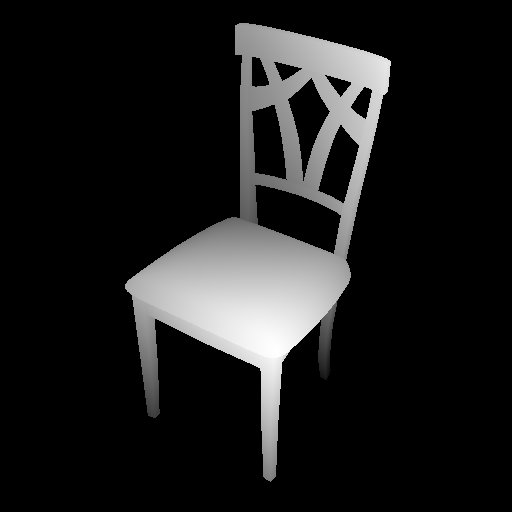}
    \end{subfigure}
    \begin{subfigure}[b]{0.09\textwidth}
        \includegraphics[width=\textwidth]{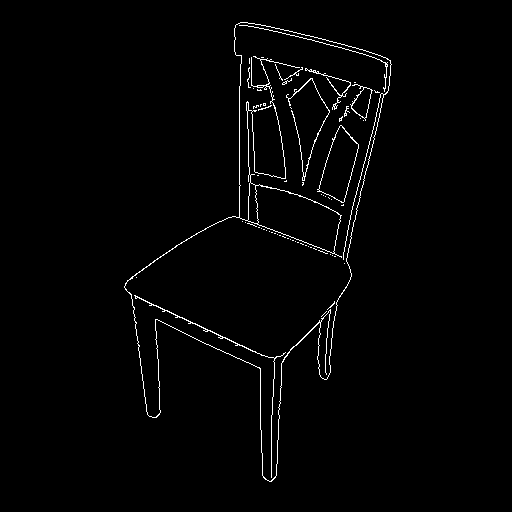}
    \end{subfigure}
    \begin{subfigure}[b]{0.09\textwidth}
        \includegraphics[width=\textwidth]{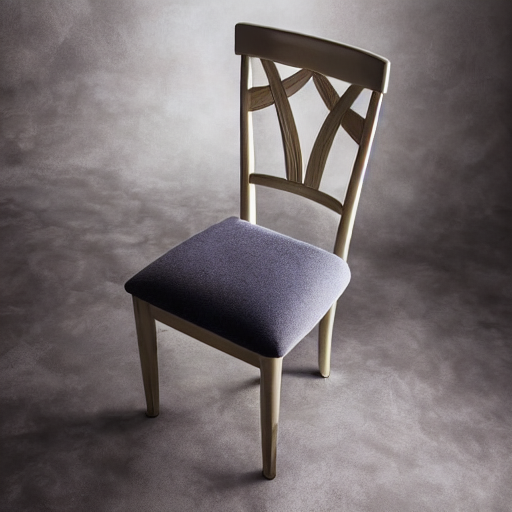}
    \end{subfigure}

    \begin{subfigure}[b]{0.09\textwidth}
        \includegraphics[width=\textwidth]{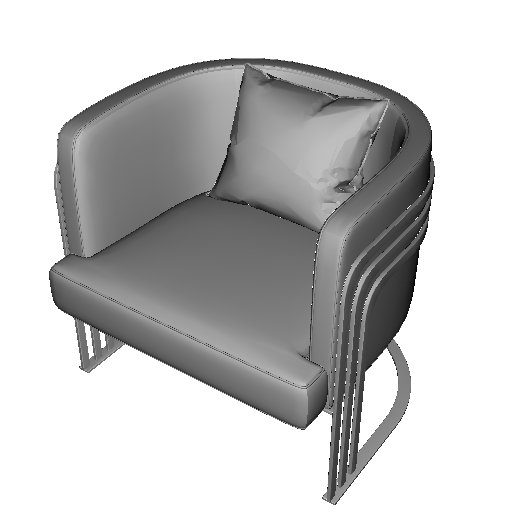}
    \end{subfigure}
    \begin{subfigure}[b]{0.09\textwidth}
        \includegraphics[width=\textwidth]{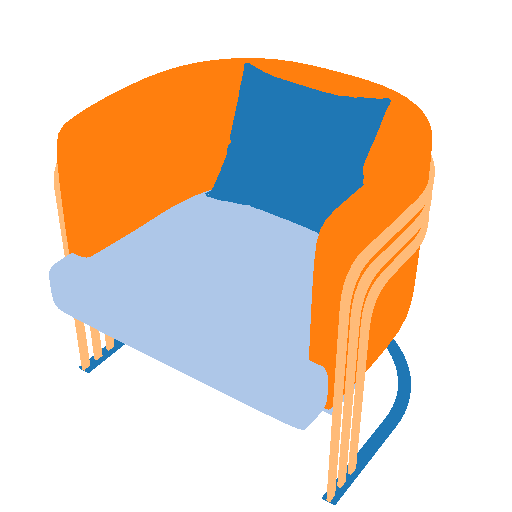}
    \end{subfigure}
    \begin{subfigure}[b]{0.09\textwidth}
        \includegraphics[width=\textwidth]{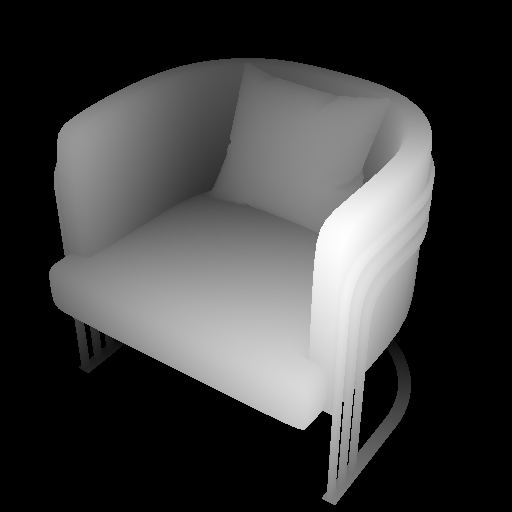}
    \end{subfigure}
    \begin{subfigure}[b]{0.09\textwidth}
        \includegraphics[width=\textwidth]{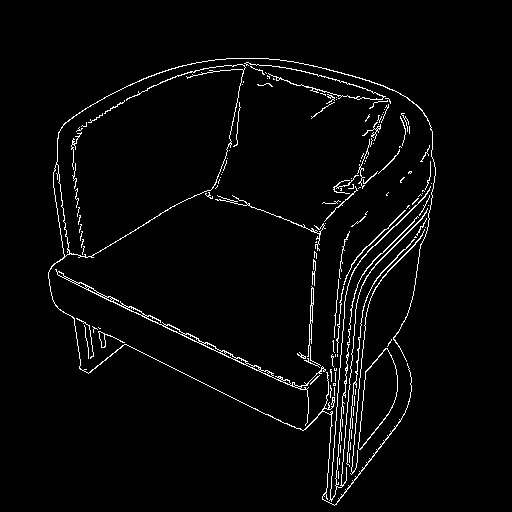}
    \end{subfigure}
    \begin{subfigure}[b]{0.09\textwidth}
        \includegraphics[width=\textwidth]{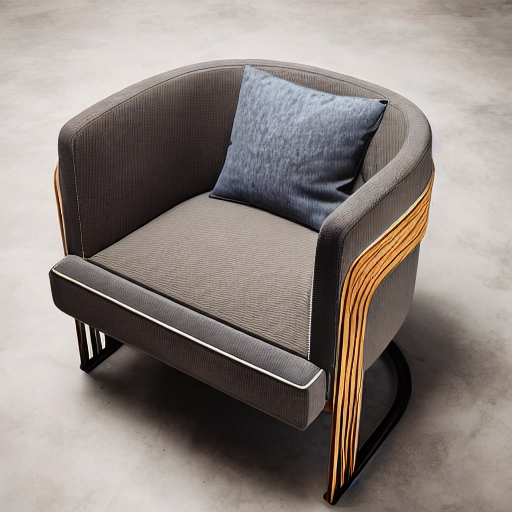}
    \end{subfigure}
  \caption{Examples (in each row from left to right) of 3D models, masks, depth images, edge images, and corresponding Stable Diffusion outputs.}
  \label{fig:model_masks}
\end{figure}

\subsection{Material Representation} \label{sec:material-representation}

To represent materials in our database, we require embeddings 
that are less sensitive to lighting, shape, and background 
context. While Alpha-CLIP~\cite{sun2023alphaclip} provides an 
effective mechanism for masking regions of interest in static 
images, it is unsuitable for representing the materials themselves. 
Alpha-CLIP embeddings retain information about the mask’s shape 
and the unmasked context, which can lead to undesirable incorporation 
of irrelevant background features when applied to material renderings.

Instead, we adopt a different preprocessing strategy for material 
representations. Since our material renderings use well-defined 
reference shapes (e.g., spheres, cones, planes, tori), we crop 
the rendered image to isolate only the textured region, ensuring 
that the embedding focuses entirely on the material. Specifically, 
we compute the largest inscribed rectangle within the region of 
interest, ensuring that the crop contains only material textures 
without any background pixels. The cropped region is then passed 
through a standard CLIP image encoder to generate a descriptor 
for each rendering (Fig. \ref{fig:overview}, left).

To account for the variation introduced by lighting conditions, 
shading, and texture distortions across different shapes, we 
compute CLIP embeddings for the full set of $n_{\text{env}} 
\times n_{\text{shapes}} = 42$ rendered images per material. 
These embeddings are arranged into a sequence and input to 
the material representation branch of our model 
(Fig.~\ref{fig:overview}, left), allowing it to learn a 
distribution of appearances that is more robust to diverse 
conditions.

\begin{table*}[t]
\caption{
Combined quantitative evaluation of all methods across four conditions: (1) main evaluation, and generalization to unseen (2) shape categories, (3) lighting, and (4) materials. We compare to variants of PhotoShape \cite{photoshape2018}, simple crop matching using CLIP \cite{radford2021_clip}, and variants of the core method of MatAtlas \cite{matatlas2024}. For CLIP and MatAtlas, accuracy is evaluated based on Top-1 (T-1) and Top-5 (T-5) metrics by taking the max. score (v1) or averaging (v2) across 42 material features. Higher accuracy values indicate better performance. \revision{Note that the values in `Unseen Materials' refer to substance-only assignment on a manually annotated dataset of LDM outputs and are therefore not directly comparable.}}
\centering
\begin{tabular}{lcccccccc}
\toprule
\multirow{2}{*}{\textbf{Method}} & 
\multicolumn{2}{c}{\textbf{Main Evaluation}} & 
\multicolumn{2}{c}{\textbf{Unseen Shapes}} & 
\multicolumn{2}{c}{\textbf{Unseen Lighting}} &
\multicolumn{2}{c}{\textbf{Unseen Materials}} \\
\cmidrule(lr){2-3} \cmidrule(lr){4-5} \cmidrule(lr){6-7} \cmidrule(lr){8-9}
& \textbf{T-1 [\%] $\uparrow$} & \textbf{T-5 [\%] $\uparrow$} 
& \textbf{T-1 [\%] $\uparrow$} & \textbf{T-5 [\%] $\uparrow$} 
& \textbf{T-1 [\%] $\uparrow$} & \textbf{T-5 [\%] $\uparrow$}
& \multicolumn{2}{c}{\textbf{T-1 [\%] $\uparrow$}} \\
\midrule
PhotoShape (ResNet-34) & 57.21 & 77.39 & 36.63 & 59.19 & 18.83 & 35.19 & \multicolumn{2}{c}{46.67} \\
PhotoShape (ResNet-50) & 58.76 & 77.62 & 38.71 & 61.42 & 20.67 & 37.31 & \multicolumn{2}{c}{49.07}  \\
PhotoShape (ViT-B/16) & 60.86 & 78.23 & 40.42 & 62.22 & 19.70 & 35.41 & \multicolumn{2}{c}{44.27}  \\
\midrule
OpenCLIP (v1) & 16.13 & 31.49 & 13.56 & 26.15 & 12.63 & 22.89 & \multicolumn{2}{c}{49.20} \\
OpenCLIP (v2) & 19.90 & 35.34 & 16.07 & 29.28 & 13.70 & 24.28 & \multicolumn{2}{c}{44.65} \\
\midrule
MatAtlas (v1) & 18.77 & 36.79 & 14.20 & 28.50 & 11.97 & 23.43 & \multicolumn{2}{c}{52.94} \\
MatAtlas (v2) & 13.14 & 28.40 & 9.03 & 20.93 & 6.62 & 14.71 & \multicolumn{2}{c}{48.93} \\
\midrule
\textbf{MatCLIP (Ours)} & \textbf{76.69} & \textbf{89.02} & \textbf{54.74} & \textbf{73.22} & \textbf{35.31} & \textbf{51.90} & \multicolumn{2}{c}{\textbf{55.47}} \\
\bottomrule
\end{tabular}
\label{tab:merged_evaluation}
\end{table*}


\subsection{MatCLIP Architecture}

Our MatCLIP model consists of two main encoders, as illustrated 
in Fig.~\ref{fig:overview} (bottom center): the \emph{Material Encoder}, denoted $E_{\text{material}}$, 
and the \emph{Part Encoder}, denoted $E_{\text{part}}$. The Material Encoder 
processes a set of rendered images of a PBR material, while the Part 
Encoder processes a masked region of an input image. The goal of 
MatCLIP is to align materials, represented by their diverse appearances 
under varying conditions, with the corresponding masked parts in input images.

The architecture is inspired by Alpha-CLIP~\cite{sun2023alphaclip}, 
which aligns text with masked portions of an image. In MatCLIP, 
we replace Alpha-CLIP's text-processing encoder with $E_{\text{material}}$, 
which processes sets of rendered PBR images. The embeddings generated by $E_{\text{material}}$ leverage the 42 
renderings of each material, as described in Section~\ref{sec:material-representation}. 
These embeddings are passed through a transformer to produce a robust 
representation of the material. In parallel, $E_{\text{part}}$ processes 
masked regions of input images, outputting embeddings that represent 
the appearance of the part in its context.

During training, embeddings from $E_{\text{material}}$ and $E_{\text{part}}$ 
are aligned using a contrastive loss, following the approach used 
in CLIP~\cite{radford2021_clip}. This loss encourages the model to 
associate corresponding embeddings while distinguishing unrelated pairs.

\subsection{Sampling Textures from Stable Diffusion}
\label{sec:sampling_textures_sd}
Generating plausible texture assignments to 3D objects is challenging because we do not know the true distribution of material assignments. LDMs offer a powerful approach to
assigning plausible materials to 3D models by capturing
realistic texture distributions from large-scale datasets.
This enables material assignments that are both coherent
and diverse. 

In our approach, a \emph{target image}, denoted $I_\text{target}$, is
a synthetic rendering produced by Stable Diffusion~\cite{rombach2021_stablediffision, Esser2024ScalingRF} (or Flux~\cite{flux2023})
that combines geometric information from a 3D model with
a stylistic interpretation of its materials. The purpose
of $I_\text{target}$ is to represent a visually plausible
appearance for the 3D model that aligns with its
structure, as shown in Fig.~\ref{fig:model_masks}~(right). 

To generate $I_\text{target}$, we condition Stable Diffusion
on depth and edge information derived from the 3D model,
ensuring that the generated textures align with the
model's geometry. Depth information is captured by
rendering a single-view depth image, \(I_\text{depth}\), from
an informative perspective, as discussed in
Section~\ref{sec:material-representation}. 
%
In addition to depth conditioning, edge information is
extracted using Canny edge detection on a diffuse gray
rendering of the 3D model. This ensures fine geometric
details, such as thin wires or fabric strips, are
preserved in \(I_\text{target}\). We also incorporate a CLIP
text embedding derived from descriptive prompts suggested by ChatGPT (e.g., “a sturdy and durable chair with a robust design”), which guides the stylistic interpretation of the material. By combining depth,
edges, and text conditioning, Stable Diffusion produces
target images that capture both geometric accuracy and
material diversity. Examples of depth images, edges,
and resulting outputs are shown in
Fig.~\ref{fig:model_masks}.

Once \(I_\text{target}\) is generated, we create part-level masks
to isolate regions for texture assignments. These masks enable precise alignment between
generated textures and specific model regions. Each part
is assigned a unique mask, as illustrated in
Fig.~\ref{fig:model_masks}.  Given a target image and masks, 
we can now use our MatCLIP model to match materials from the MatSynth dataset. The whole procedure of using Stable Diffusion together with our model is shown in Fig.~\ref{fig:overview} (top center).

\section{Results}\label{sec:results}

We evaluate our method across three key 
dimensions: (1) material classification performance, 
demonstrating MatCLIP’s effectiveness in identifying 
materials across diverse shapes and lighting conditions; 
(2) generalization to new inputs, including manually 
annotated evaluations of substances derived from latent 
diffusion model outputs; and (3) ablation studies assessing 
the impact of dataset diversity on classification accuracy. 
Complementary qualitative evaluations (Section~\ref{sec:ldm-qualitative}) 
illustrate MatCLIP's ability to extend beyond known material 
distributions. Together, these experiments validate MatCLIP’s 
robustness in procedural modeling tasks and its superiority 
over state-of-the-art methods.

\revision{
We base our model on a pre-trained ViT-B/16 version of Alpha-CLIP. In particular, our part encoder $E_{part}$ is based on its image encoder, taking an RGBA image as input. For the material encoder $E_{material}$, we leverage Alpha-CLIP's fine-tuned transformer-based text encoder and use a linear layer to map to the transformer’s input dimension. For the training procedure, we use a batch size of 256 and a learning rate of 1.0e-4. 
}
Our model is trained on a GPU cluster node with 4 NVIDIA 
A100 40GB cards for approximately 48 hours. Hyperparameters 
follow the contrastive loss strategy introduced in 
Alpha-CLIP~\cite{sun2023alphaclip}. MatCLIP adapts this approach 
to align part-level embeddings with diverse material representations, 
as detailed in Section~\ref{sec:material-representation}. This 
enables MatCLIP to robustly handle variations in shape and lighting 
conditions while maintaining consistent performance.

For quantitative evaluations, we partition the dataset at 
the object level to ensure no object appears in both training 
and test sets, even when rendered with different material 
assignments. This design prevents model overfitting to specific 
shapes and ensures unbiased evaluations. While all materials 
are included in both training and test sets, the object-level 
separation highlights MatCLIP’s ability to generalize across 
new shape configurations within a closed material set.

Qualitative evaluations presented in Section~\ref{sec:ldm-qualitative} 
further demonstrate MatCLIP’s ability to generalize to novel 
material distributions. These results highlight MatCLIP’s 
potential to extend beyond quantitative metrics, reinforcing 
its applicability to real-world procedural modeling tasks.

\subsection{Classification Performance}
\label{sec:classification}
We evaluate MatCLIP's material classification performance by 
ranking materials in the test split of our dataset, using 
descriptors generated by the part encoder (\( E_{\text{part}} \)) 
and material encoder (\( E_{\text{material}} \)). Materials 
are ranked by cosine similarity between their embeddings. 
For each query shape from the test set, materials were matched 
from the entire set of available materials, without restricting 
classification to the substance categories defined for the 
3DCoMPaT++ shapes. This ensured that the model was evaluated 
across the full database of materials. The Top-1 and Top-5 
accuracies of our method, compared to PhotoShape~\cite{photoshape2018} 
and MatAtlas~\cite{matatlas2024}, are shown in 
Table~\ref{tab:merged_evaluation} (Main Evaluation). 

For a fair comparison, we modified PhotoShape by replacing 
its original ResNet34 and ResNet50 backbones with the more 
recent ViT-B/16 architecture to better align with modern 
methods. This resulted in slightly improved performance of 
PhotoShape on our data. Additionally, we evaluated MatAtlas 
without restricting classification to the material classes 
suggested by ChatGPT, instead of testing its ability to match 
materials from the full database. Although this adjustment 
reduced MatAtlas's performance, we intended to assess only 
the classification component, ensuring the comparison focused 
on embedding quality rather than substance category 
limitations. These adjustments aimed to standardize the 
evaluation setup and ensure all methods were tested under 
similar conditions. Despite these efforts to strengthen the 
baselines, MatCLIP achieves the highest accuracy, with a 
Top-1 accuracy of 76.69\% and a Top-5 accuracy of 89.02\%, 
significantly outperforming the next-best method, PhotoShape 
(ViT-B/16) which had a top-1 accuracy of 60.86\%.

These results underscore the robustness of \( E_{\text{part}} \) 
and \( E_{\text{material}} \) in generating descriptors that 
effectively handle the complexity of diverse objects and 
material assignments, establishing MatCLIP as a state-of-the-art 
approach for material classification.

\revision{
\subsection{Generalization to Unseen Shape Categories}
\label{sec:generalization_shapes}
While our network was trained only on renderings of chairs from the 3DCoMPaT++ shape dataset \cite{li20223d_compat, slim_3dcompatplus_2023}, it is not restricted to chairs but generalizes well to arbitrary shape categories. For our experiments, we rendered 3007 images across all non-chair shape categories of 3DCoMPaT++. We removed very bad renderings showing heavy over- or underexposure to finally obtain 2997 renderings and corresponding 11k part-mask pairs to run the actual experiments. As shown in Table \ref{tab:merged_evaluation} (Unseen Shapes), our results on unseen shape categories show a slight performance drop, our method however still performs significantly better than the baseline methods. While we used only chairs for training for this proof of concept, we believe that the performance of our method could be improved by incorporating a larger variety of different shape categories into the training procedure.
}


\revision{
\subsection{Generalization to Unseen Lighting Conditions}
\label{sec:generalization_lights}
We focus on studio lighting conditions to train our network because we argue that this best reflects the lighting conditions of product and shape images generated by latent diffusion models (LDMs) like Stable Diffusion \cite{rombach2021_stablediffision} and Flux \cite{flux2023}. Still, our method is able to generalize well to unseen and also more extreme (e.g. colored) lighting as we show in Table \ref{tab:merged_evaluation} (Unseen Lighting). For our experiment, we download 769 HDRI environment maps from all available categories (e.g. outside, inside, daylight, artificial lighting, colored lighting) from Polyhaven and render 4502 images. Again, we remove outliers showing heavy over- or underexposure to obtain 4475 images and 18k part-mask pairs. While the performance of our model both on the Top-1 and Top-5 accuracy is lower than on unseen shapes, it still performs significantly better than the baseline methods we compare to. We believe that there is still room for improvement by incorporating a larger number of different lighting scenarios (e.g. natural lighting captured at different day times, colored lighting, single- and multi-source lights) into our training dataset.}


\subsection{Generalization to Unseen Materials}
To evaluate MatCLIP’s performance in assigning 
materials not explicitly part of the training set, we implemented 
a GUI for manual annotation of outputs from Stable Diffusion. This was needed because the ground truth was not known for LDM outputs. 
It was used to assign the most likely substance to 
highlighted shape parts. In total, approximately 400 annotations 
were collected, covering various parts from 100 different 
3DCoMPaT++ chair models.

We then compared MatCLIP’s Top-1 ranked material 
assignments to those of competing methods, including 
PhotoShape~\cite{photoshape2018} and MatAtlas~\cite{matatlas2024}. 
The results are summarized in Table~\ref{tab:merged_evaluation} (Unseen Materials). 
Although MatCLIP consistently outperforms other approaches, the 
performance gap is narrower than in prior evaluations. This is 
likely due to the inherent ambiguity and subjectivity in manual 
annotations.

For a fair comparison, the same baseline configurations were 
used as in Section~\ref{sec:classification}. Notably, we evaluated 
each method’s ability to rank substances from the full material 
database, without restricting classification to substance 
categories predefined in 3DCoMPaT++. Despite this broader 
evaluation scope, MatCLIP achieves a Top-1 accuracy of 55.47\%, 
outperforming all competitors.

\subsection{Ablation Study}

To assess the impact of dataset diversity on material classification 
accuracy, we conducted an ablation study by varying the number of shapes 
and environment maps used to render materials. Our full model 
relies on 42 renderings per material, generated using 6 distinct shapes 
and 7 HDR environment maps. We systematically reduced the dataset by 
testing subsets, including cases with a single shape across all environment 
maps and all shapes under a single environment map. For each subset, we retrained MatCLIP and 
monitored Top-1 classification accuracy over the first 9,000 training steps.


The results  highlight the significant 
advantage of using the full dataset (cf. Table~\ref{tab:ablation}). The full model achieves the highest 
accuracy, significantly outperforming models trained on subsets. Notably, 
the worst performance was observed with a single shape under a single 
environment map, emphasizing the importance of both 
geometric and lighting diversity for learning robust material representations.


These findings confirm that diverse shapes and lighting conditions are 
crucial for MatCLIP’s performance. Future improvements could leverage an 
even broader set of base shapes and environment maps to further enhance 
material generalization and classification robustness.

\begin{table}[t]
\caption{Ablation study examining the effect of shape and environment map diversity on classification accuracy. The full model, which uses all 6 shapes and 7 environment maps (42 renderings per material), performs best.}
\centering
\begin{tabular}{lc}
\toprule
\textbf{Model Name} & \textbf{Val. Acc. [\%] $\uparrow$} \\
\midrule
Plane Shape, 1 Environment Map  &  7.01 \\
Plane Shape, 3 Environment Maps & 49.53 \\
Plane Shape, 5 Environment Maps & 52.47 \\
Plane Shape, All Environment Maps & 53.19 \\
\midrule
3 Shapes, 1 Environment Map & 52.24 \\
5 Shapes, 1 Environment Map & 54.54 \\
All Shapes, 1 Environment Map  & 56.68 \\
\midrule
\textbf{MatCLIP (Ours), Full Model} & \textbf{60.54} \\
\bottomrule
\end{tabular}
\label{tab:ablation}
\end{table}



\subsection{Qualitative Evaluations with Latent~Diffusion~Models}\label{sec:ldm-qualitative}

We conduct qualitative evaluations using outputs from Stable Diffusion (cf. Figure~\ref{fig:results_ours}) and Flux (cf. Figures~\ref{fig:results_flux} and \ref{fig:results_main_flux_mix}) to assess MatCLIP's ability to generalize beyond the closed set of training materials. These evaluations focus on assigning materials to synthetic target images ($I_\text{target}$) generated using depth, edge, and text conditioning, as described in Section~\ref{sec:material-representation}. Our model selects the best-matching material from the MatSynth database for each target image, producing visually coherent and contextually appropriate assignments, even for complex shapes and lighting conditions.

Figures~\ref{fig:results_ours}, \ref{fig:results_flux}, and \ref{fig:results_main_flux_mix} showcase examples where MatCLIP robustly matches materials to parts of a 3D shape. The visual results illustrate the effectiveness of our embeddings in achieving plausible and realistic material assignments across diverse target distributions. \revision{Please refer to the supplementary materials for more results.}

\subsection{Prompt-Based Design Exploration}
\label{sec:chatgpt}
To explore a larger design space, we leverage the power of ChatGPT to generate a large number of prompts. In general, we distinguish between two types. We use a \emph{regular prompt} that usually only consists of a few words like "a comfortable chair." The prompts we try to obtain from ChatGPT we refer to as \emph{advanced prompts}, since they usually consist of very detailed descriptions of materials and the relationship between the materials that are used for different parts of a shape. To allow ChatGPT to describe such relationships, we do not only use a textual prompt but also provide an exemplar image of the shape of interest. Using advanced prompts, we are able to obtain a larger variety than we were able to get using only regular prompts (cf. Figures 1 and 4 in the supplementaries).


\subsection{User Study}
To evaluate the effectiveness of our method for assigning plausible materials to large non-textured shape collections, we conducted a user study on MTurk. Specifically, we selected 100 chair shapes at random from 3DCoMPaT++ and used Stable Diffusion, as detailed in Sec.~\ref{sec:sampling_textures_sd}, to generate plausible material assignments. For each chair shape, we created approximately five different sets of material assignments, resulting in a total of 506 textured chair shapes.


To benchmark our approach, we compared it to the procedural material assignment method from 3DCoMPaT++, which procedurally assigns materials to shape parts based on predefined substance categories. All chair shapes, including those textured using our method and the procedural approach, were rendered using Blender. To ensure a fair comparison, we used identical rendering parameters for all models, including the same camera settings (intrinsic and extrinsic), environment map, and overall scene composition.


We constructed a dataset of image pairs, each consisting of one chair textured using our method and one textured using the procedural approach. The order of images within each pair was randomized, and participants on MTurk were asked to choose the chair they believed had the more plausible material assignment. We collected two independent responses (HITs) per image pair to increase reliability, resulting in 1,012 ranked decisions. 

The results demonstrate a clear preference for our method: in 783 of the 1,012 pairs (77.37\%), participants selected the chair textured using our approach as having more plausible material assignments. In contrast, the procedural method from 3DCoMPaT++ was preferred in only 229 pairs (22.63\%). These findings underscore the effectiveness of our approach, leveraging the material distributions suggested by Stable Diffusion, in producing more realistic and contextually appropriate textures for large shape collections compared to existing procedural methods.

\section{Limitations and Future Work}

\begin{figure}[t] 
    \centering
    \begin{subfigure}[b]{0.45\linewidth}
        \includegraphics[width=\linewidth]{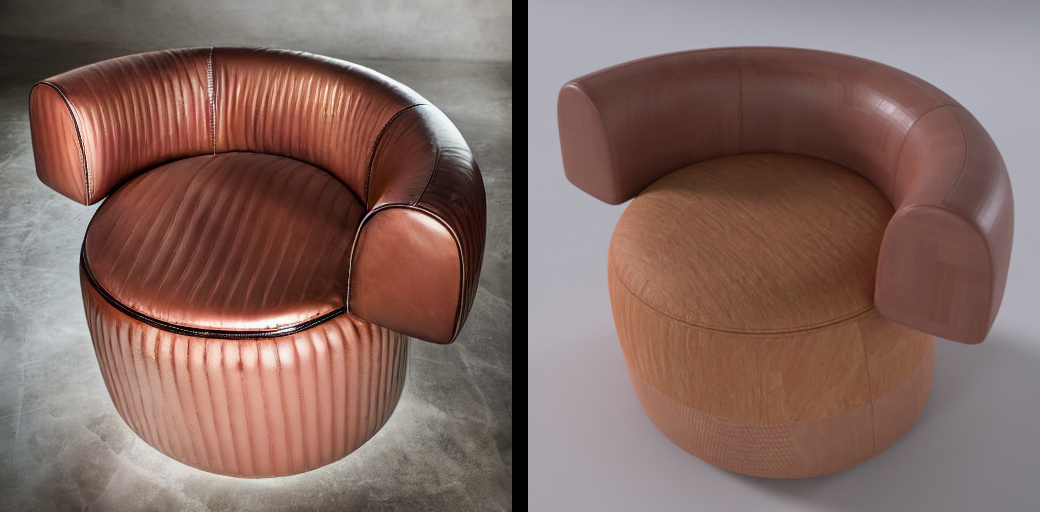}
    \end{subfigure}
    \begin{subfigure}[b]{0.45\linewidth}
        \includegraphics[width=\linewidth]{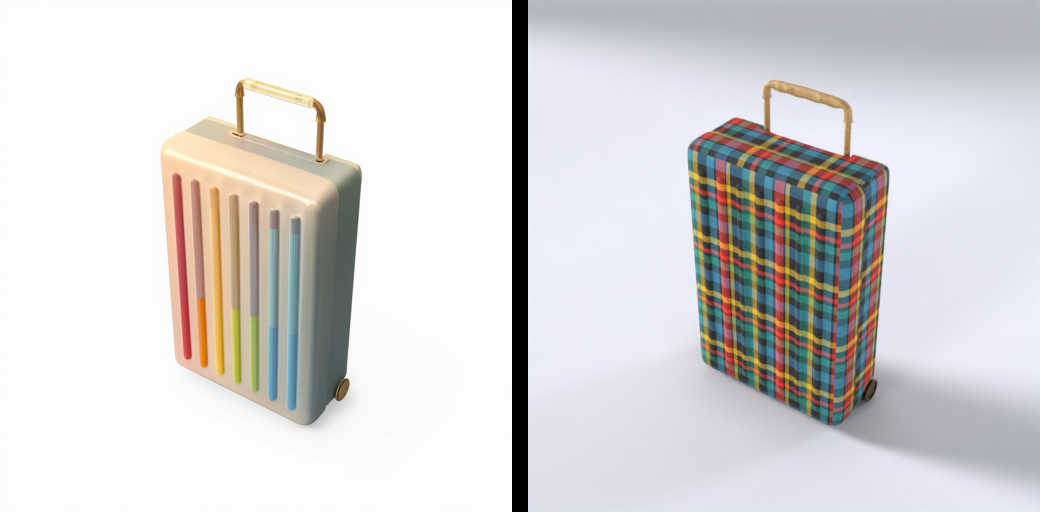} 
    \end{subfigure}
    \caption{MatCLIP struggles when Stable Diffusion (left side) or Flux (right side) add additional geometry (e.g., wrinkles on the left side) or when no material from the material database would give a good match (right side).}
    \label{fig:limitations}
\end{figure}

Despite MatCLIP's strong performance in assigning materials 
to 3D shape parts, our method has certain limitations. One 
challenge arises when Stable Diffusion introduces geometric 
artifacts, such as fine-structured geometric details not present in 
the original shape, despite conditioning on depth and edge 
inputs (cf. Fig.~\ref{fig:limitations}). These artifacts 
can occasionally lead to material assignments that appear 
inconsistent with the input.

Generalization to unseen material distributions remains an 
open challenge. While qualitative evaluations (cf. 
Section~\ref{sec:ldm-qualitative}) suggest that MatCLIP can 
produce visually coherent material assignments for unseen 
scenarios, quantitative evaluations using manually annotated 
substances show room for improvement. The Top-1 accuracy 
drops from 76.69\% on the validation set of known materials 
(cf. Table~\ref{tab:merged_evaluation} (Main Evaluation)) to 55.47\% for manual 
annotations (cf. Table~\ref{tab:merged_evaluation}, (Unseen Materials)). However, 
this reduction is not unique to our method; other methods 
also experience performance drops under similar conditions. 
We believe this is mainly due to the uncertainty in the ground truth annotations, as the task of material assignment from RGB images is also very difficult for human labelers.

\revision{In the future, we want to investigate potential alternative use cases like using our model in classifier-guided diffusion-based material optimization instead of just retrieval.} 

\section{Conclusion}\label{sec:conclusion}






\revision{
This paper introduced MatCLIP, a framework for robust material classification and assignment to 3D objects. By training on diverse renderings, MatCLIP learns embeddings invariant to shape and lighting, achieving a Top-1 accuracy of 76.69\% on MatSynth and significantly outperforming prior methods.

Beyond classification, MatCLIP supports plausible material assignments to 3D shapes using LDM-generated target images, making a connection between procedural modeling and high-quality PBR materials. By relying on public resources like MatSynth and 3DCoMPaT++, our approach ensures reproducibility and accessibility for the community.
}

\begin{acks}
  We thank the anonymous reviewers for their constructive comments. The work is supported by funding from KAUST - Center of Excellence for Generative AI, under award number 5940, and the NTGC-AI program.
\end{acks}


\bibliographystyle{ACM-Reference-Format}

\bibliography{bibliography}

\newcommand{\imgwidth}{0.135}
\newcommand{\imgskip}{0.5mm}
\newcommand{\tupleskip}{4mm}

\begin{figure*}
  \centering
    
    
    \begin{tabular}{@{}c@{\hskip \imgskip}c@{\hskip \imgskip}c@{\hspace{\tupleskip}}c@{\hskip \imgskip}c@{\hskip \imgskip}c@{}} 
        Stable Diffusion & Ours & 3DCoMPaT++ & Stable Diffusion & Ours & 3DCoMPaT++ \\ 
        [2mm]
        \includegraphics[width=\imgwidth\textwidth]{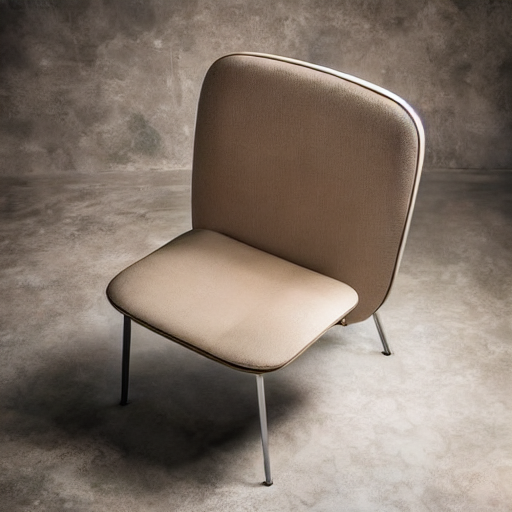} &
        \includegraphics[width=\imgwidth\textwidth]{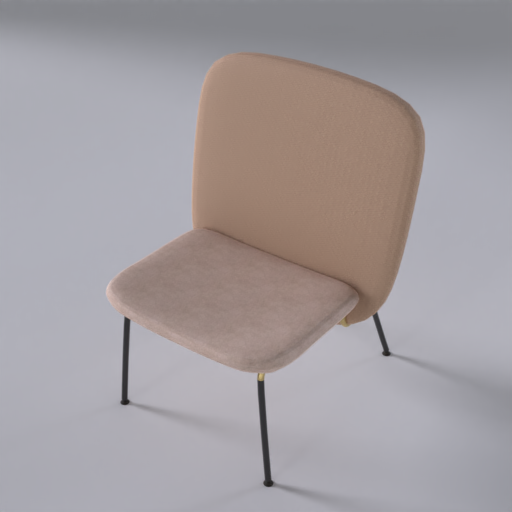} &
        \includegraphics[width=\imgwidth\textwidth]{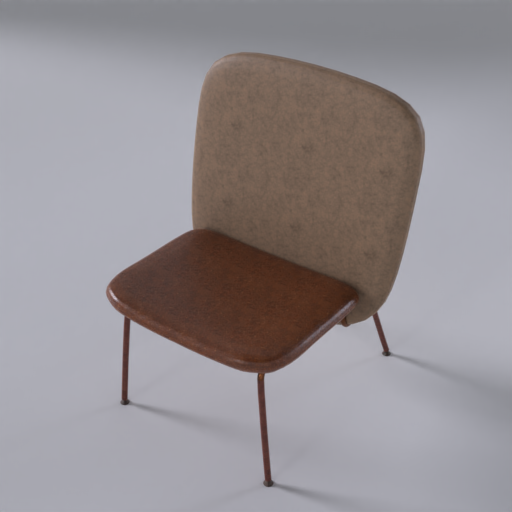} &
        \includegraphics[width=\imgwidth\textwidth]{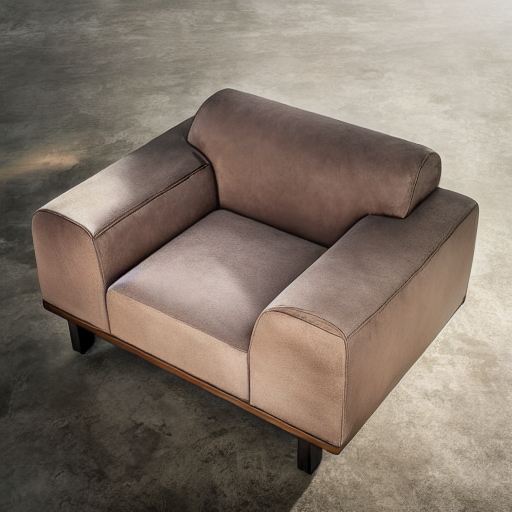} &
        \includegraphics[width=\imgwidth\textwidth]{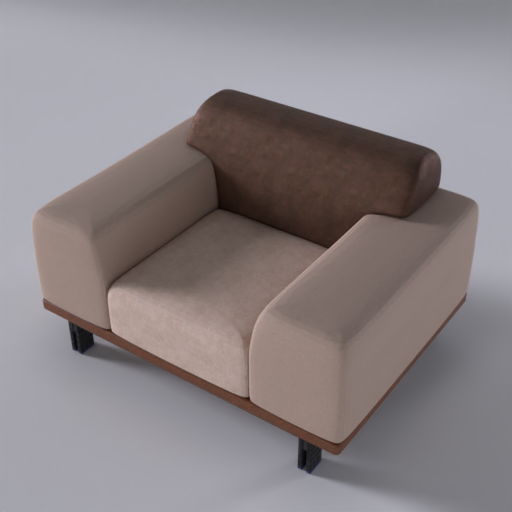} &
        \includegraphics[width=\imgwidth\textwidth]{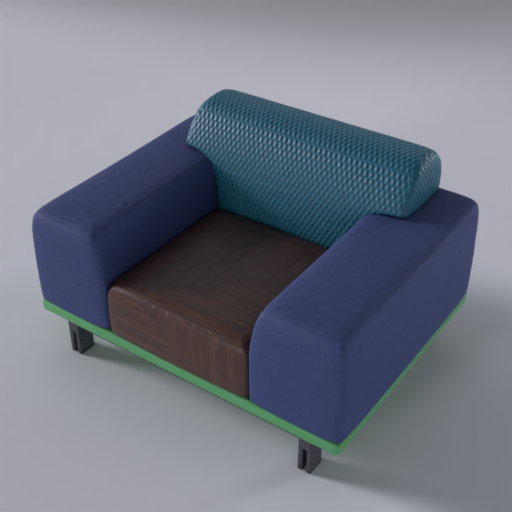} \\
        [\tupleskip]
        \includegraphics[width=\imgwidth\textwidth]{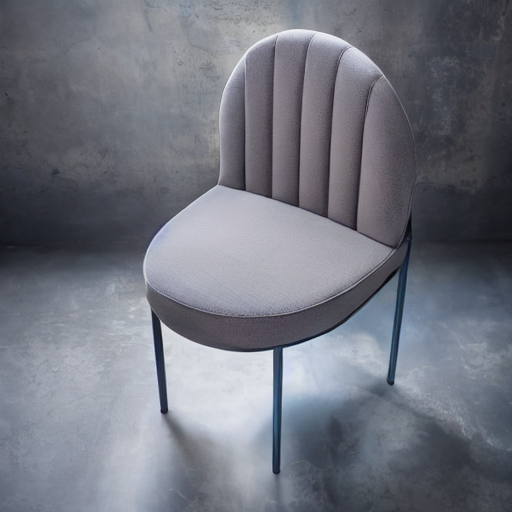} &
        \includegraphics[width=\imgwidth\textwidth]{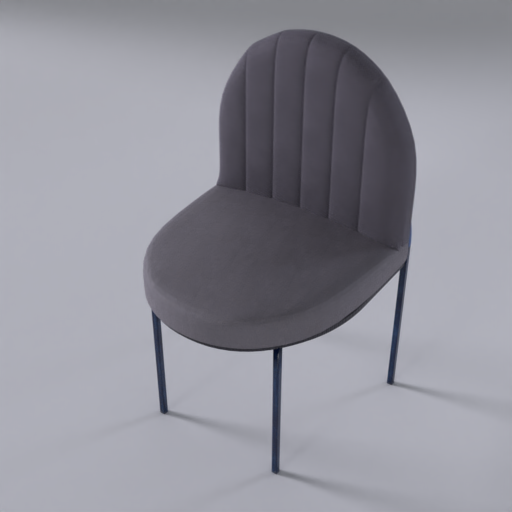} &
        \includegraphics[width=\imgwidth\textwidth]{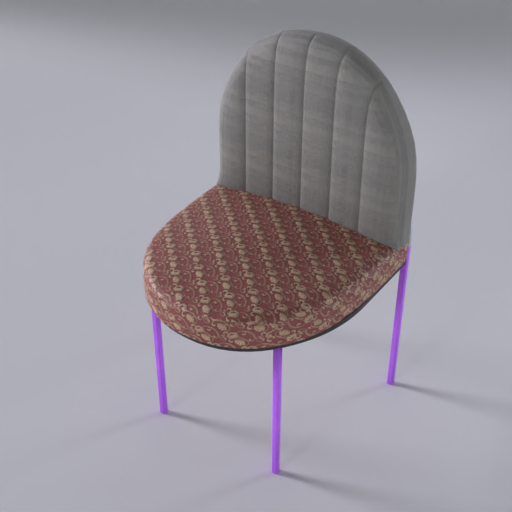} &
        \includegraphics[width=\imgwidth\textwidth]{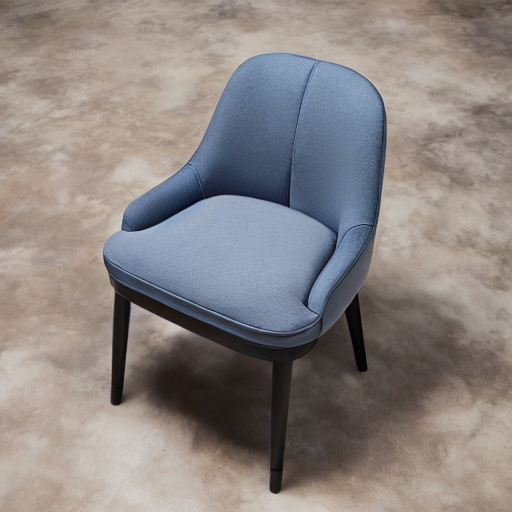} &
        \includegraphics[width=\imgwidth\textwidth]{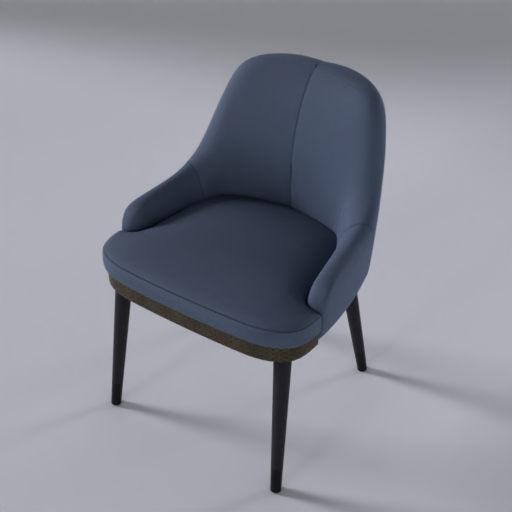} &
        \includegraphics[width=\imgwidth\textwidth]{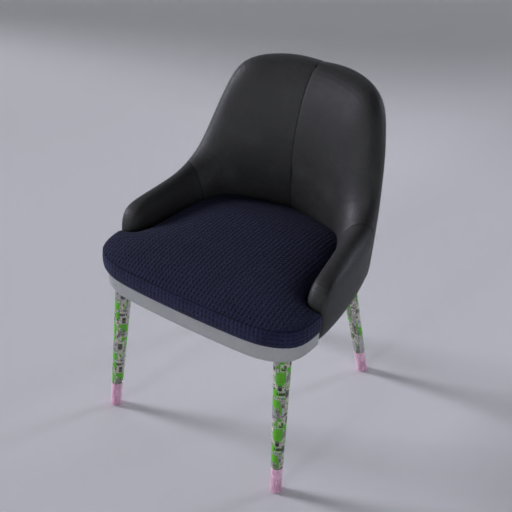} \\
        [\tupleskip]
        \includegraphics[width=\imgwidth\textwidth]{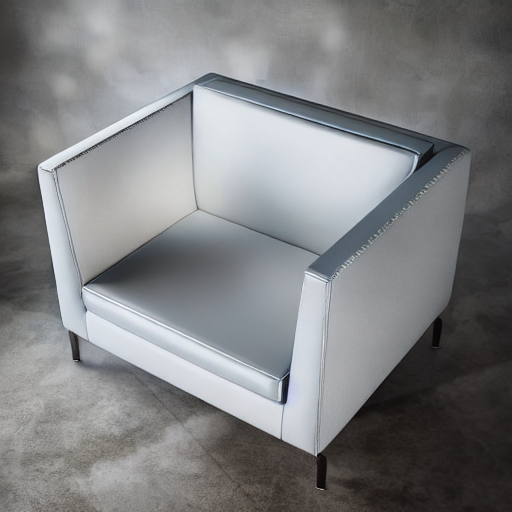} &
        \includegraphics[width=\imgwidth\textwidth]{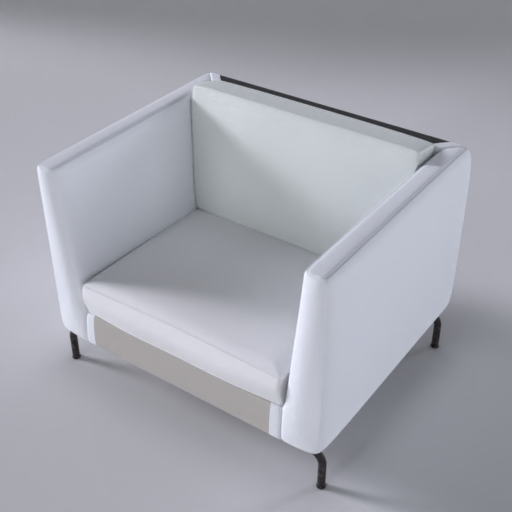} &
        \includegraphics[width=\imgwidth\textwidth]{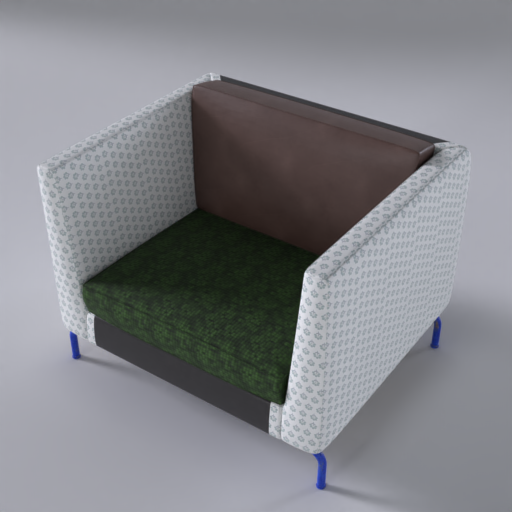} &
        \includegraphics[width=\imgwidth\textwidth]{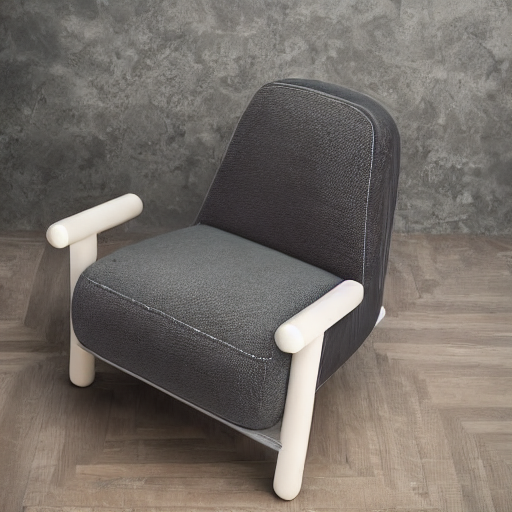} &
        \includegraphics[width=\imgwidth\textwidth]{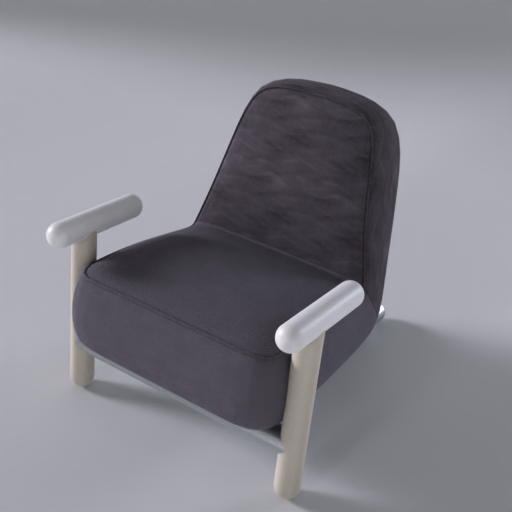} &
        \includegraphics[width=\imgwidth\textwidth]{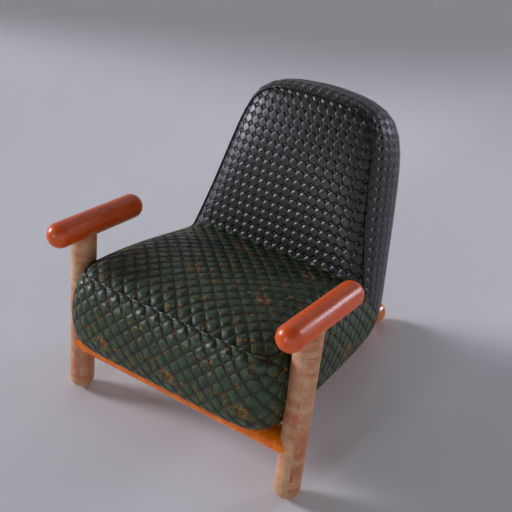} \\
        [\tupleskip]
        \includegraphics[width=\imgwidth\textwidth]{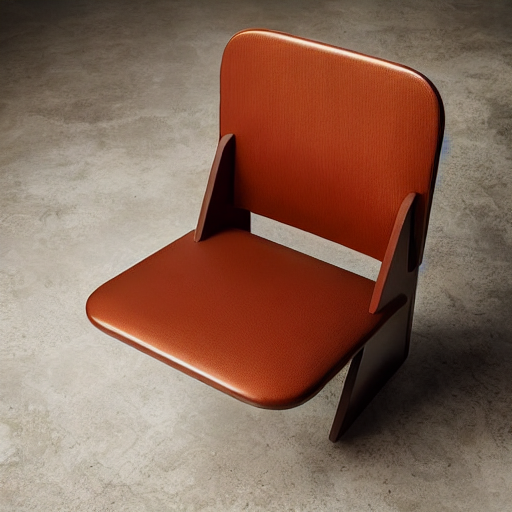} &
        \includegraphics[width=\imgwidth\textwidth]{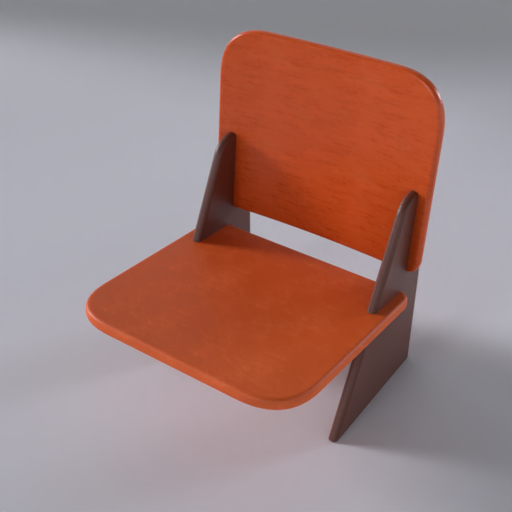} &
        \includegraphics[width=\imgwidth\textwidth]{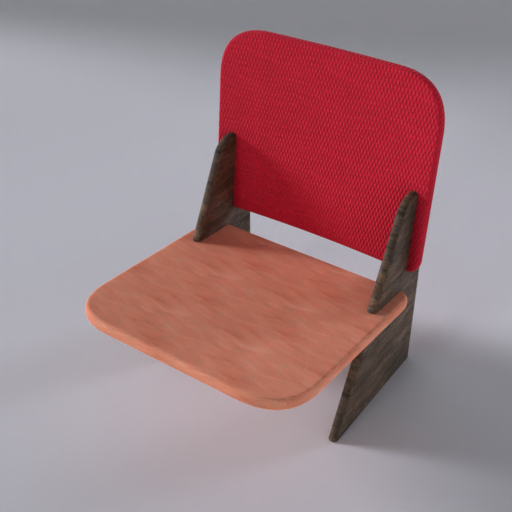} &
        \includegraphics[width=\imgwidth\textwidth]{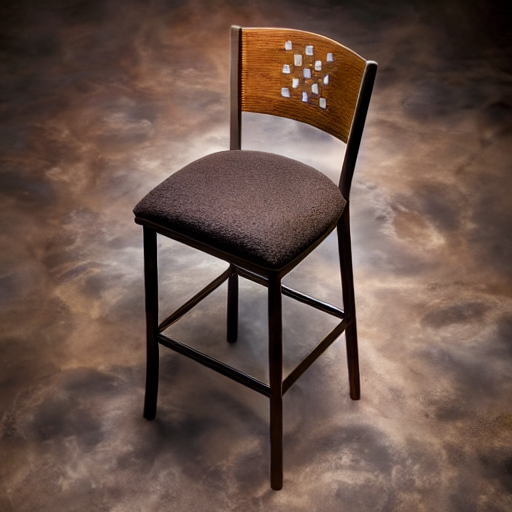} &
        \includegraphics[width=\imgwidth\textwidth]{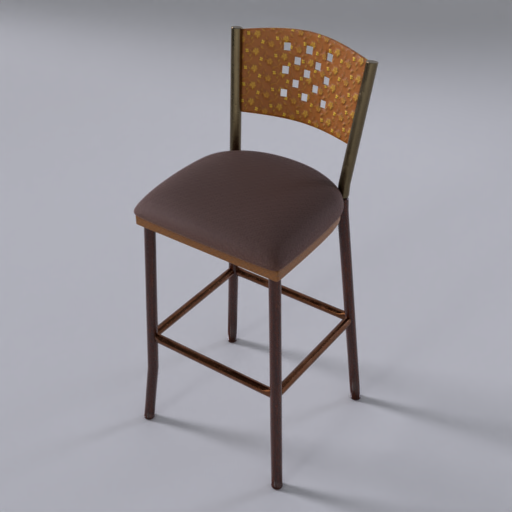} &
        \includegraphics[width=\imgwidth\textwidth]{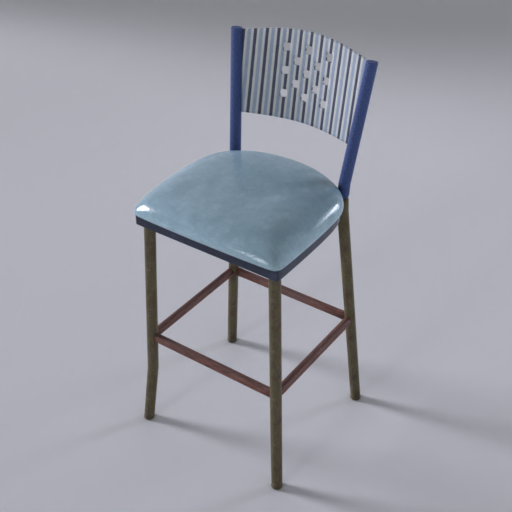} \\
        [\tupleskip]
        \includegraphics[width=\imgwidth\textwidth]{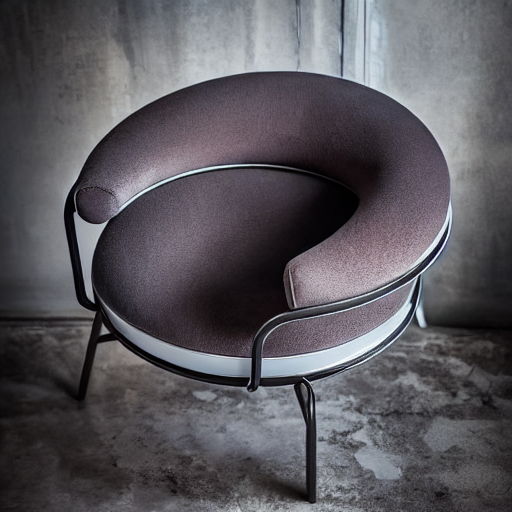} &
        \includegraphics[width=\imgwidth\textwidth]{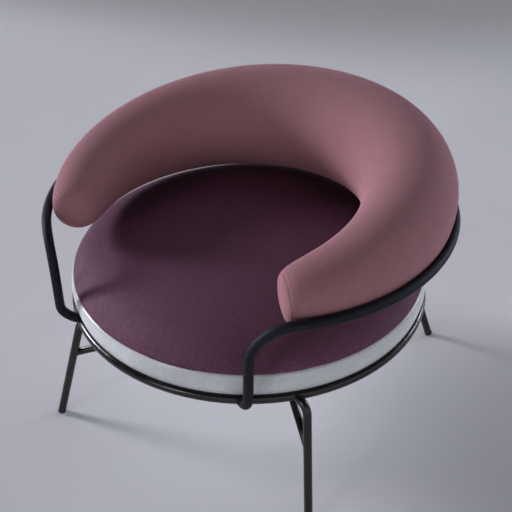} &
        \includegraphics[width=\imgwidth\textwidth]{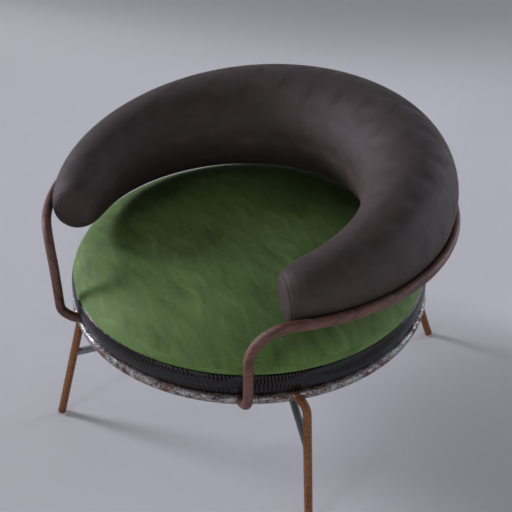} &
        \includegraphics[width=\imgwidth\textwidth]{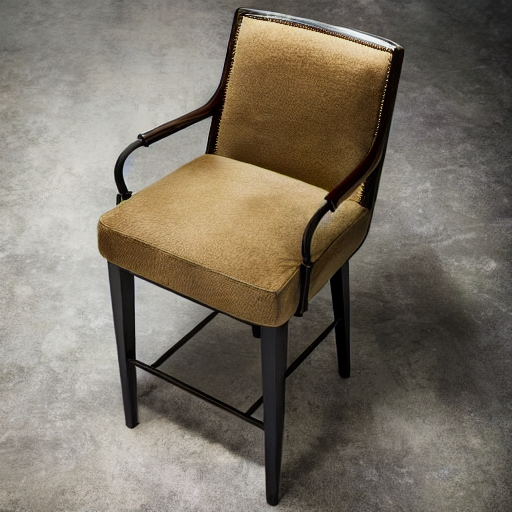} &
        \includegraphics[width=\imgwidth\textwidth]{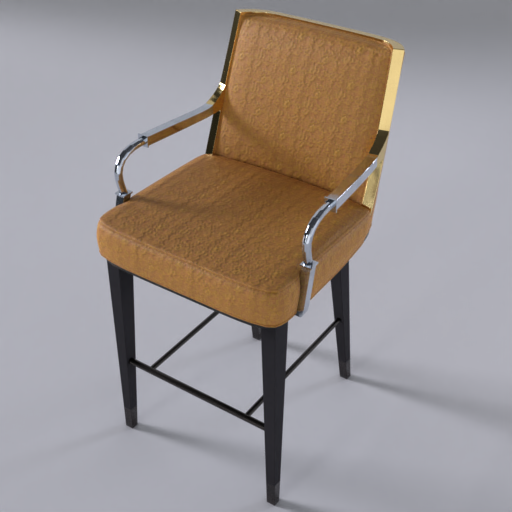} &
        \includegraphics[width=\imgwidth\textwidth]{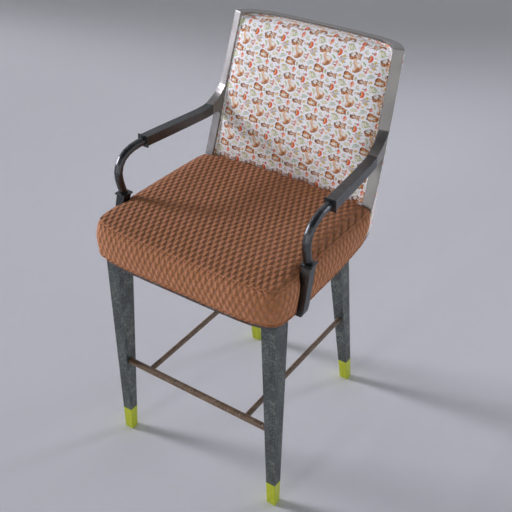} \\
        [\tupleskip]
        \includegraphics[width=\imgwidth\textwidth]{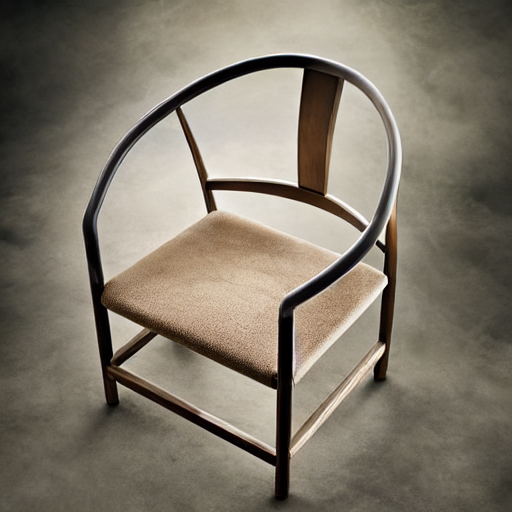} &
        \includegraphics[width=\imgwidth\textwidth]{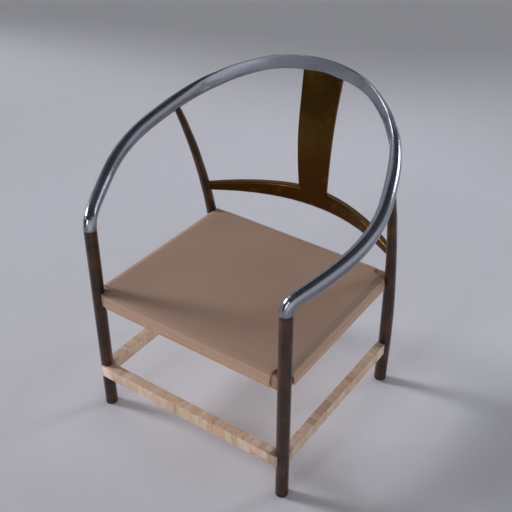} &
        \includegraphics[width=\imgwidth\textwidth]{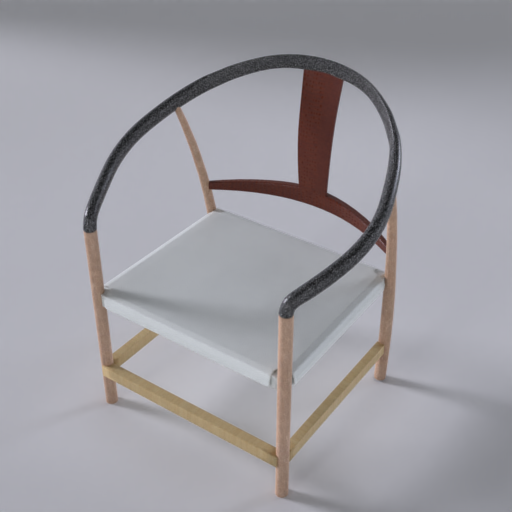} &
        \includegraphics[width=\imgwidth\textwidth]{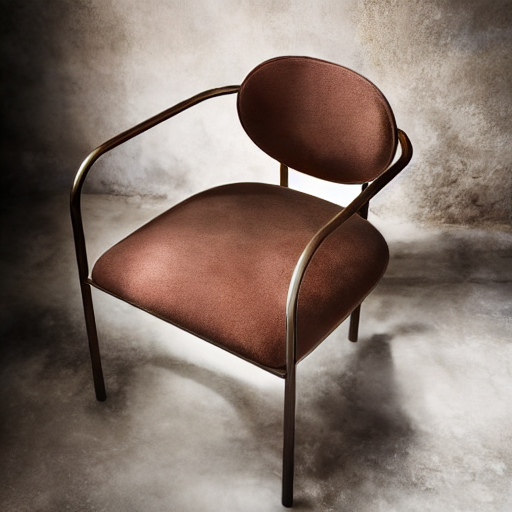} &
        \includegraphics[width=\imgwidth\textwidth]{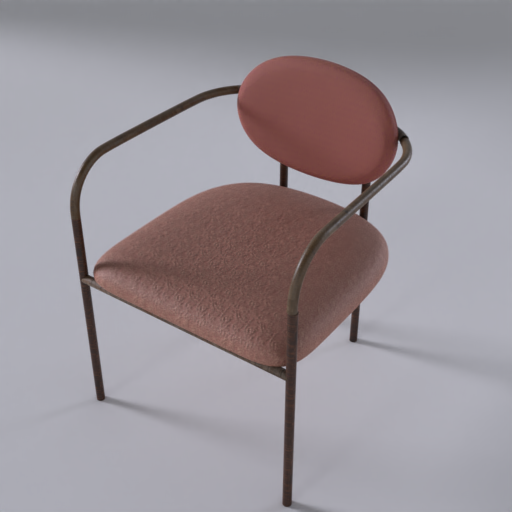} &
        \includegraphics[width=\imgwidth\textwidth]{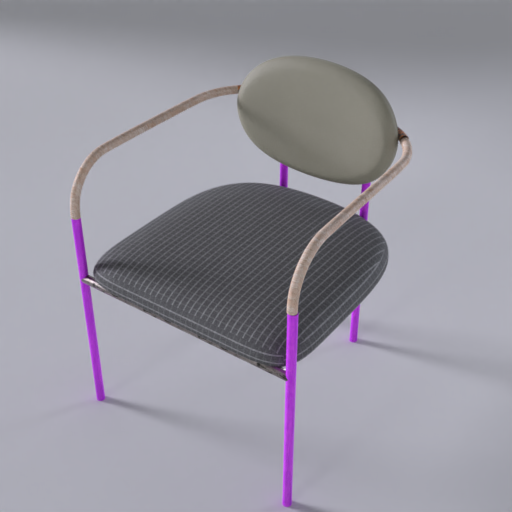} \\
        [\tupleskip]
        \includegraphics[width=\imgwidth\textwidth]{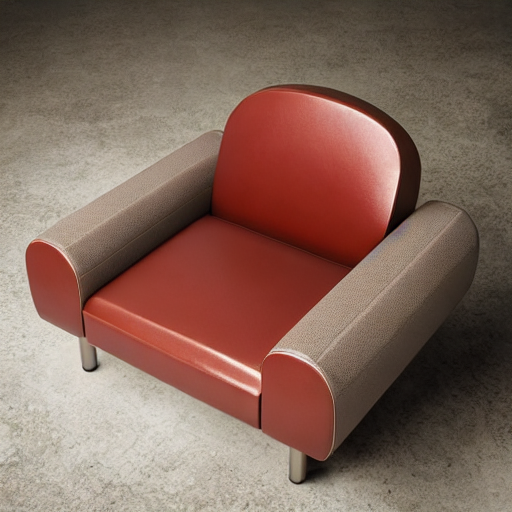} &
        \includegraphics[width=\imgwidth\textwidth]{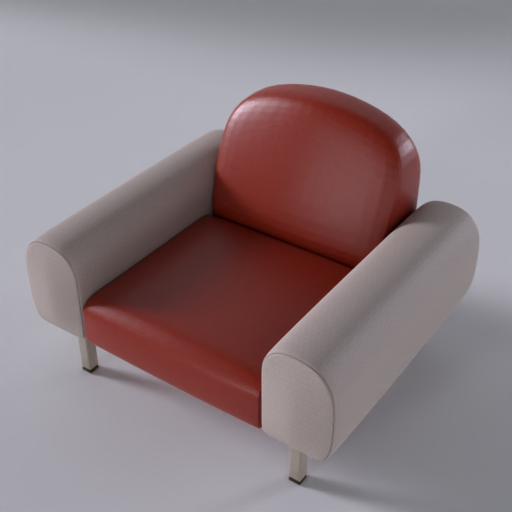} &
        \includegraphics[width=\imgwidth\textwidth]{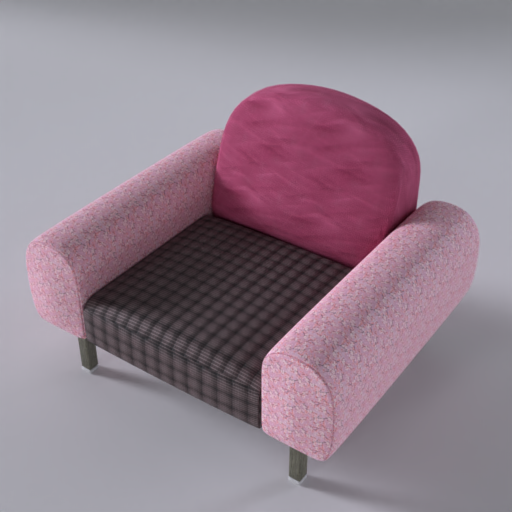} &
        \includegraphics[width=\imgwidth\textwidth]{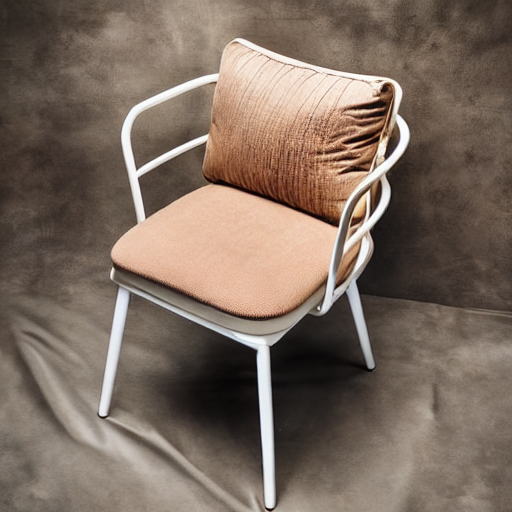} &
        \includegraphics[width=\imgwidth\textwidth]{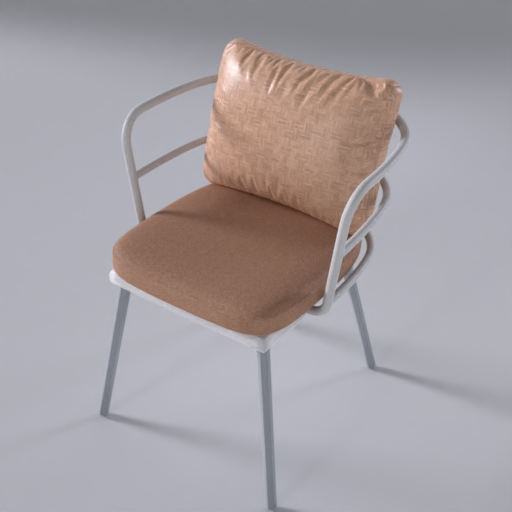} &
        \includegraphics[width=\imgwidth\textwidth]{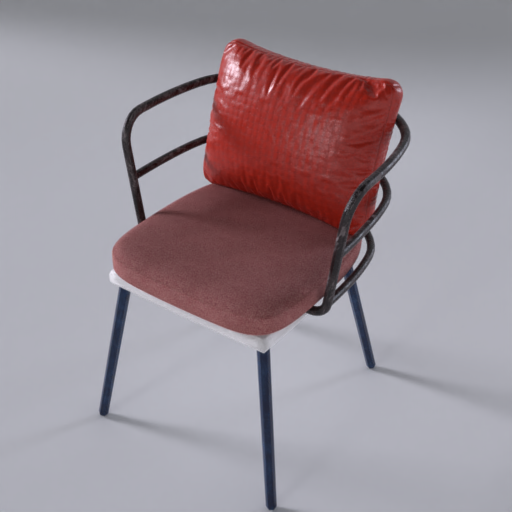}

        
    \end{tabular}
    \caption{Matching materials from Stable Diffusion output images to materials of our material database using our MatCLIP model. Each 3-tuple shows the output from Stable Diffusion (left side), a rendering of the same shape with the materials matched by our MatCLIP (center), and a procedural assignment from suitable substance categories by 3DCoMPaT++ \cite{li20223d_compat, slim_3dcompatplus_2023}.}
    \label{fig:results_ours}
\end{figure*}

\renewcommand{\imgwidth}{0.1325}
\renewcommand{\imgskip}{0.5mm}
\renewcommand{\tupleskip}{4mm}

\newcommand{\imageflux}[1]{\includegraphics[width=\imgwidth\textwidth, trim={0 0 538px 0}, clip]{#1}}
\newcommand{\imageours}[1]{\includegraphics[width=\imgwidth\textwidth, trim={538px 0 0 0}, clip]{#1}}

\begin{figure*}
  \centering
    \begin{tabular}{@{}c@{\hskip \imgskip}c@{\hspace{\tupleskip}}c@{\hskip \imgskip}c@{\hspace{\tupleskip}}c@{\hskip \imgskip}c@{}}
        Flux & MatCLIP & Flux & MatCLIP & Flux & MatCLIP \\
        \imageflux{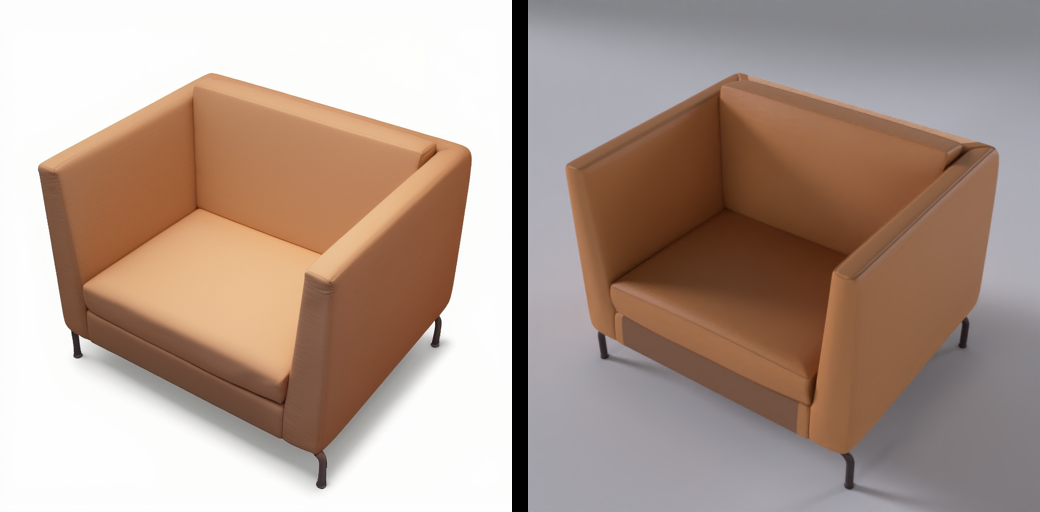} &
        \imageours{figs/results_flux/0c_0ad_00008.png} &
        \imageflux{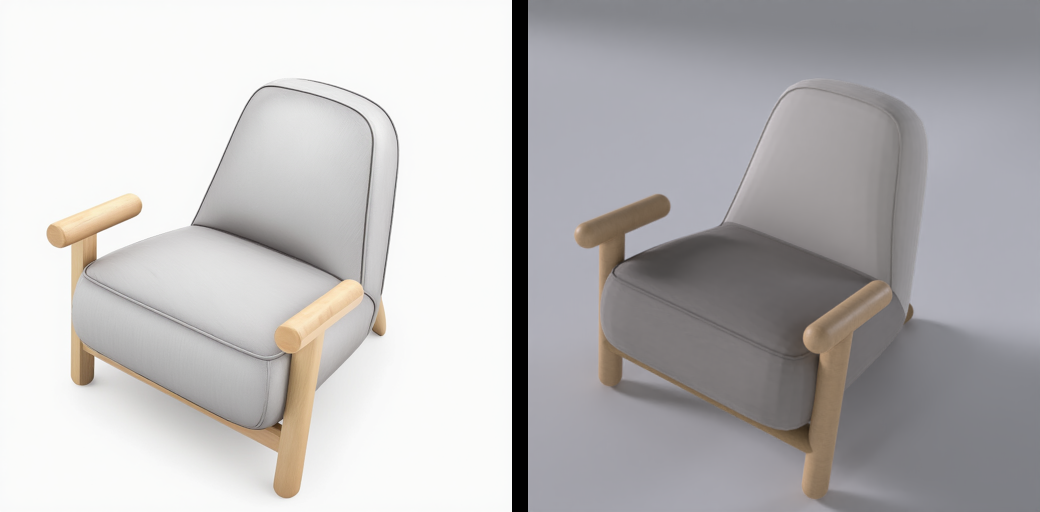} &
        \imageours{figs/results_flux/0c_0b8_00033.png} &
        \imageflux{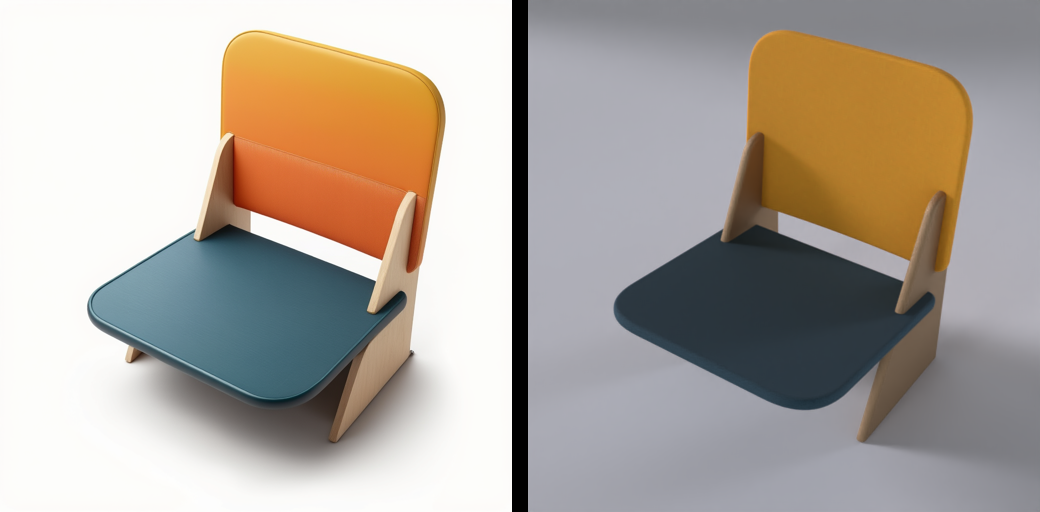} &
        \imageours{figs/results_flux/0c_0c4_00022.png} \\
        \imageflux{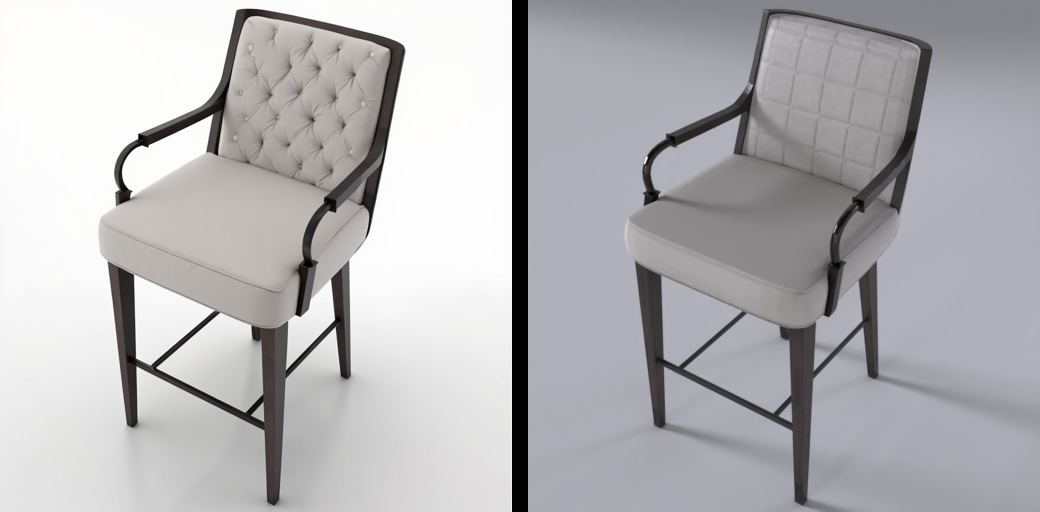} &
        \imageours{figs/results_flux/0c_0fd_00047.png} &
        \imageflux{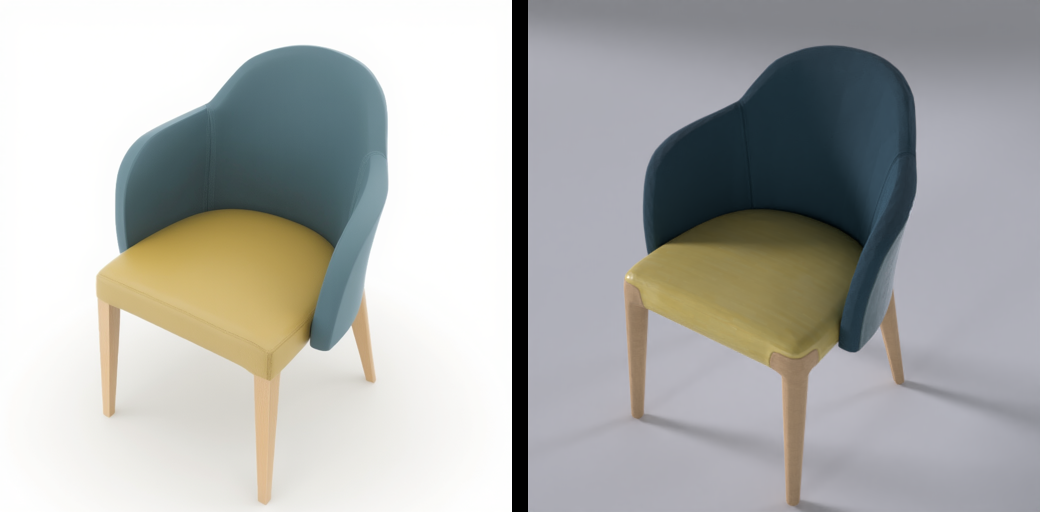} & 
        \imageours{figs/results_flux/0c_359_00039.png} & 
        \imageflux{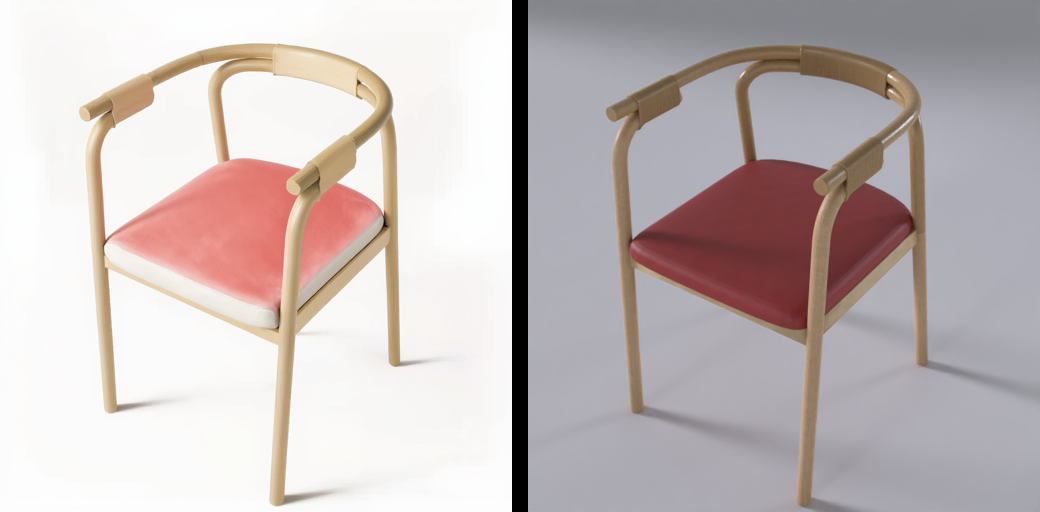} &
        \imageours{figs/results_flux/0c_2ae_00058.png} \\
        \imageflux{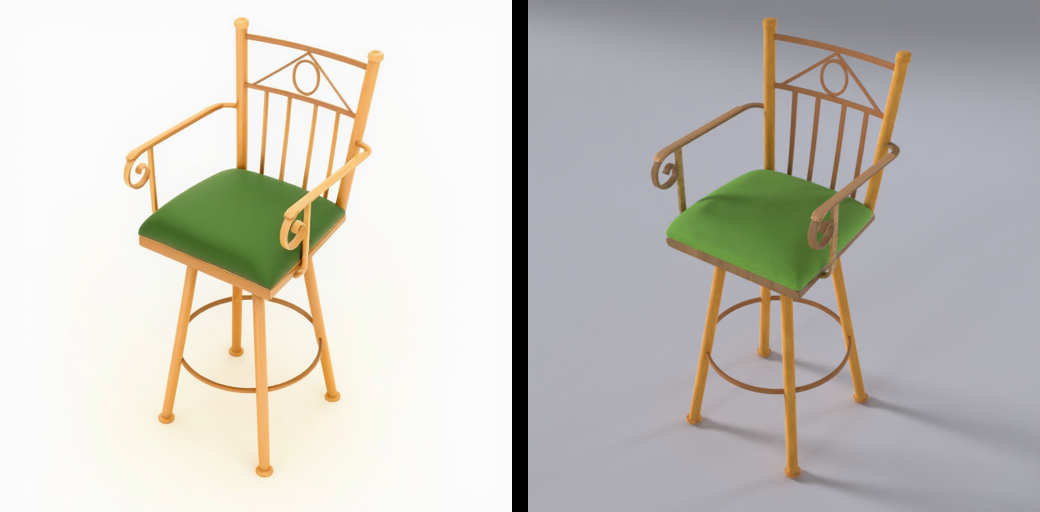} &
        \imageours{figs/results_flux/0c_3bc_00027.png} &
        \imageflux{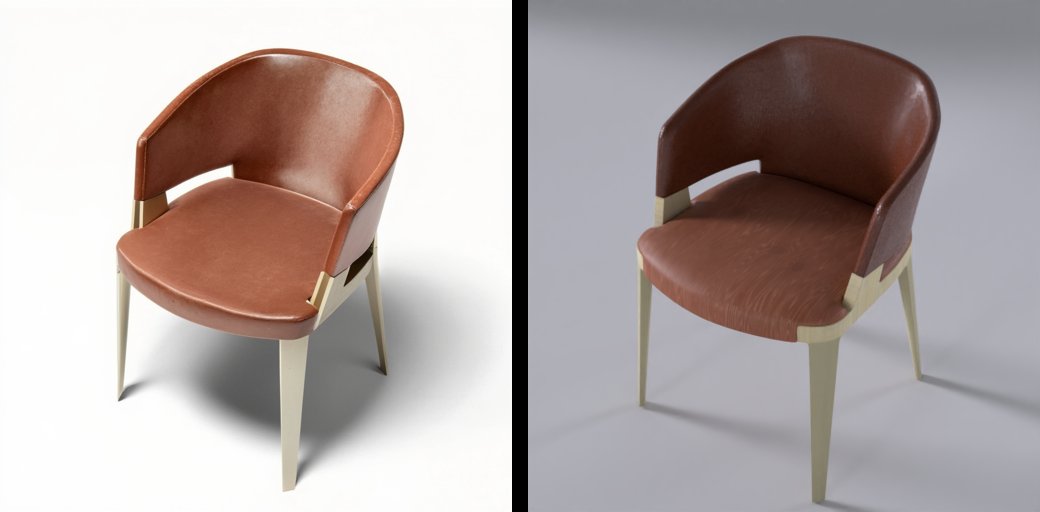} &
        \imageours{figs/results_flux/0c_42c_00042.png} &
        \imageflux{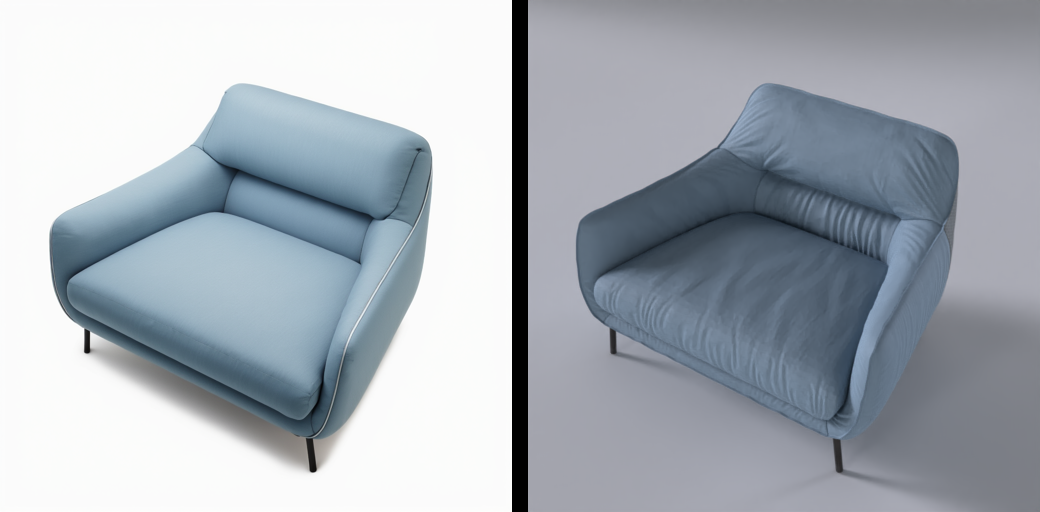} &
        \imageours{figs/results_flux/0c_214_00001.png} \\
        \imageflux{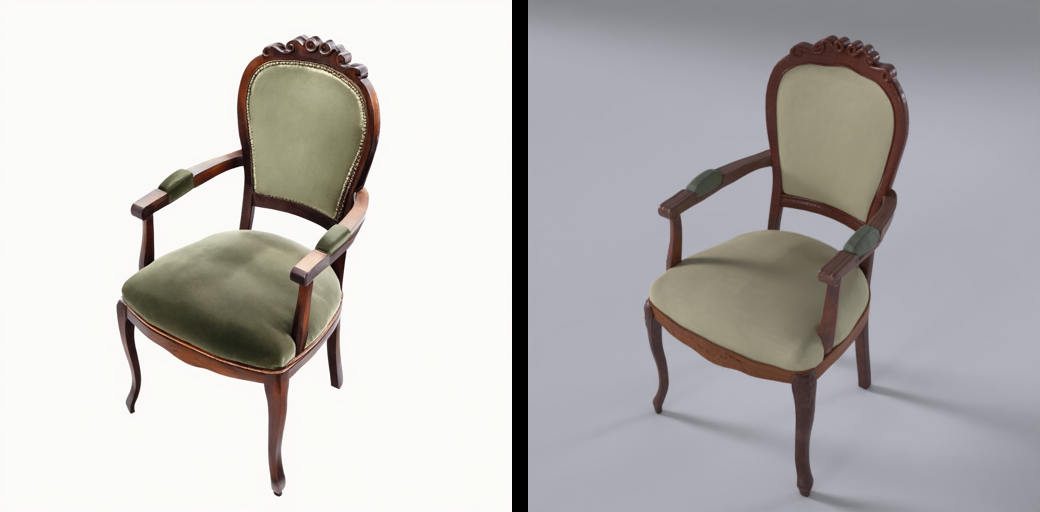} &
        \imageours{figs/results_flux/0c_187_00021.png} &
        \imageflux{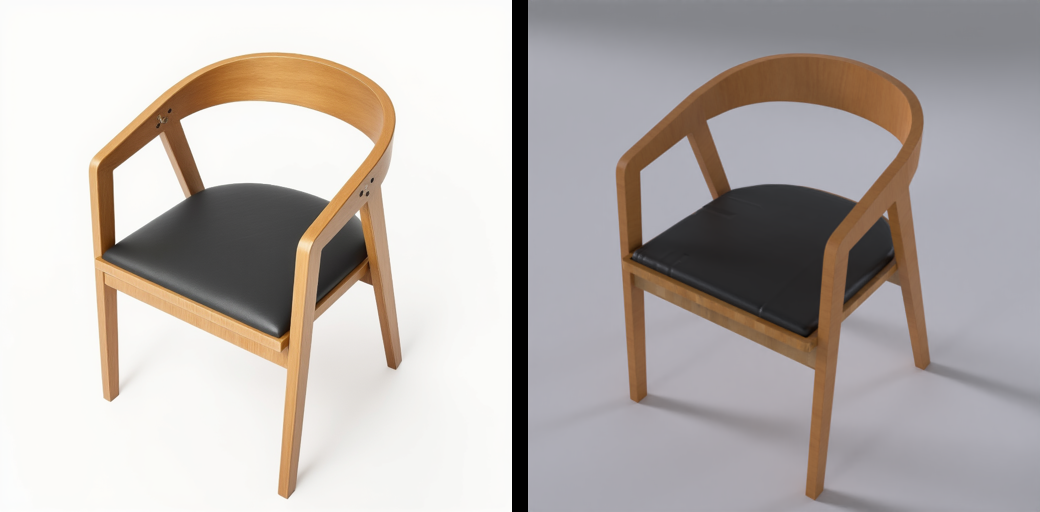} &
        \imageours{figs/results_flux/0c_396_00038.png} &
        \imageflux{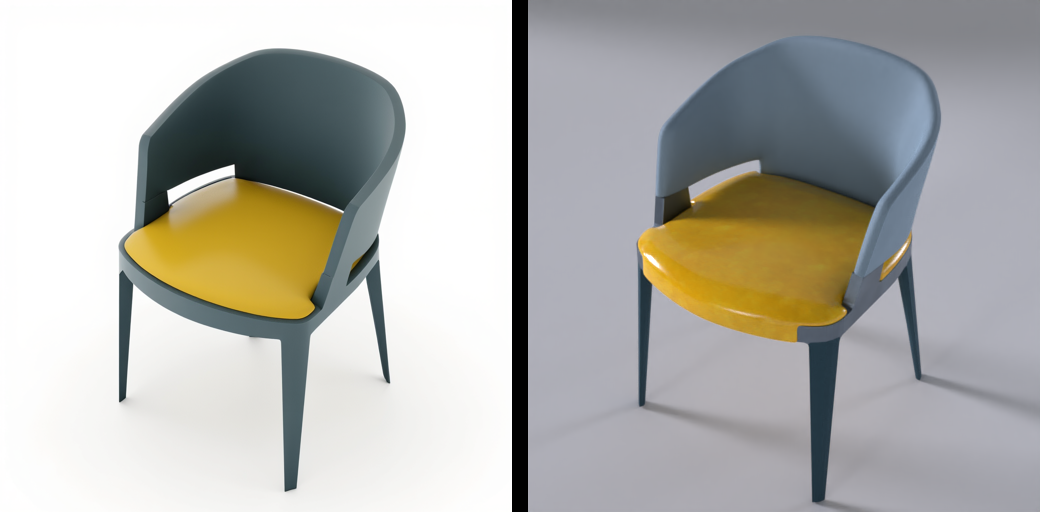} &
        \imageours{figs/results_flux/0c_42c_00039.png}
        
    \end{tabular}
    \caption{Material matching results using MatCLIP on chair images generated by Flux \cite{flux2023}.}
    \label{fig:results_flux} 
\end{figure*}




\renewcommand{\imageflux}[1]{\includegraphics[width=\imgwidth\textwidth]{#1}}
\renewcommand{\imageours}[1]{\includegraphics[width=\imgwidth\textwidth]{#1}}

\begin{figure*}
	\centering
	\begin{tabular}{@{}c@{\hskip \imgskip}c@{\hspace{\tupleskip}}c@{\hskip \imgskip}c@{\hspace{\tupleskip}}c@{\hskip \imgskip}c@{}}
		Flux & MatCLIP & Flux & MatCLIP & Flux & MatCLIP \\
		\imageflux{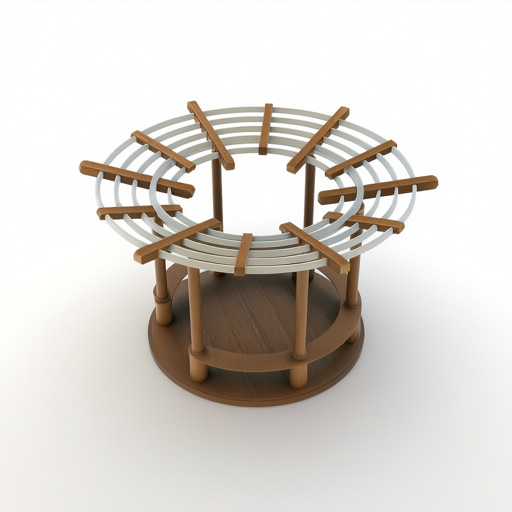} &
        \imageours{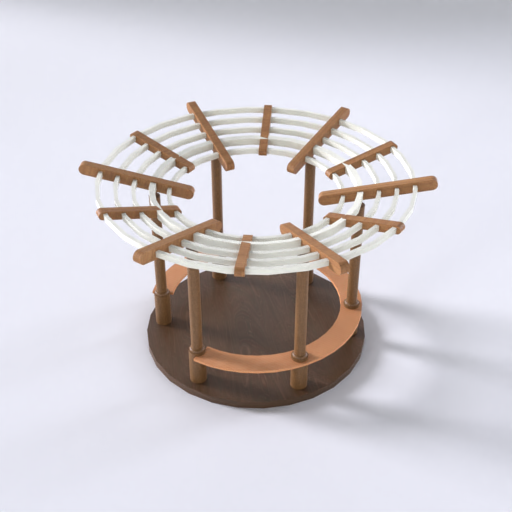} &
        \imageflux{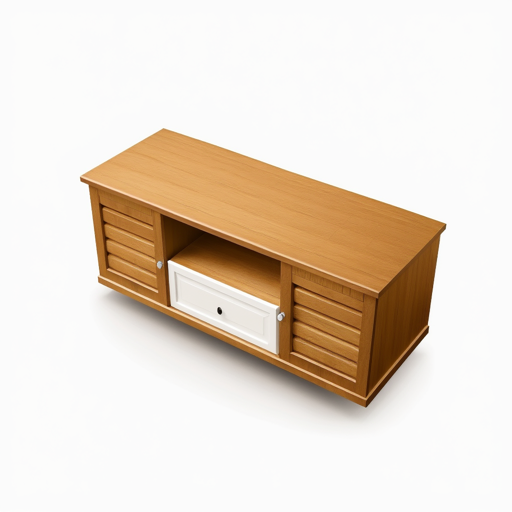} &
        \imageours{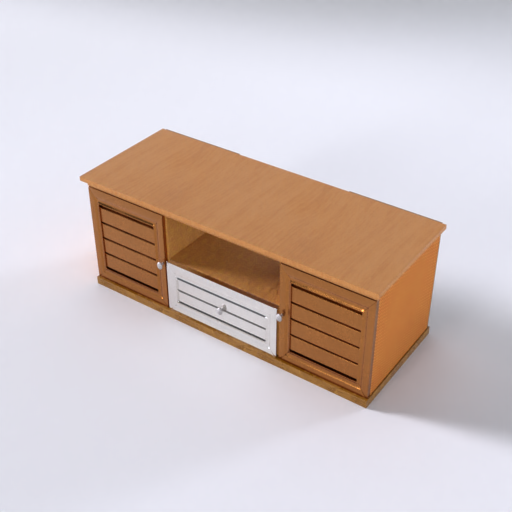} &
        \imageflux{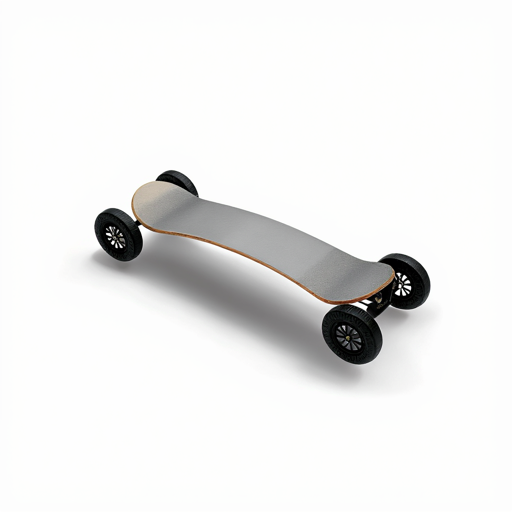} &
        \imageours{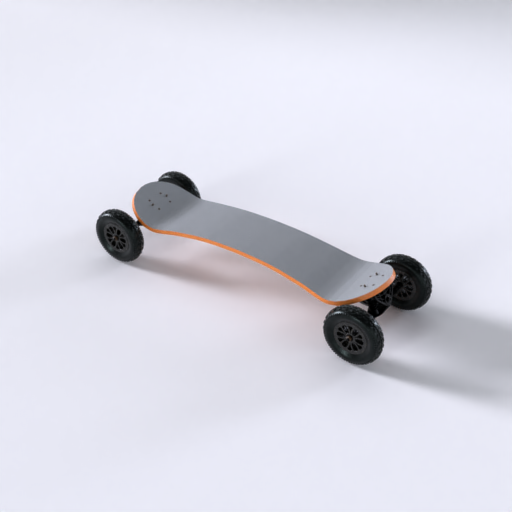} \\
        \imageflux{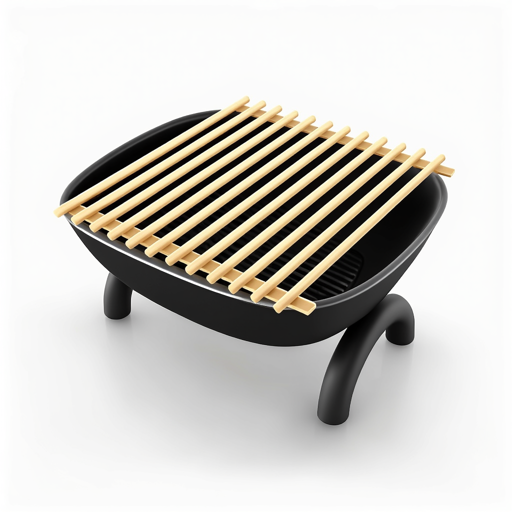} &
        \imageours{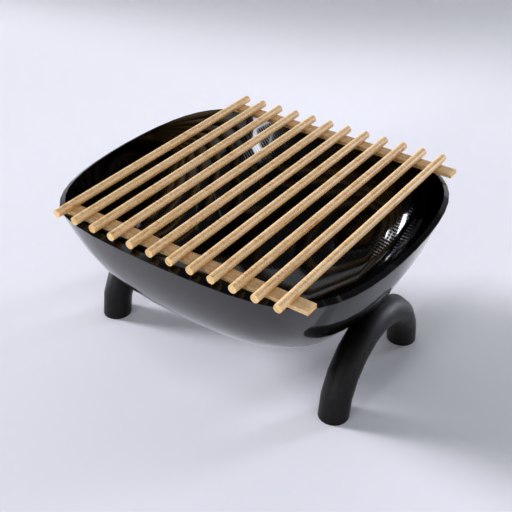} &
        \imageflux{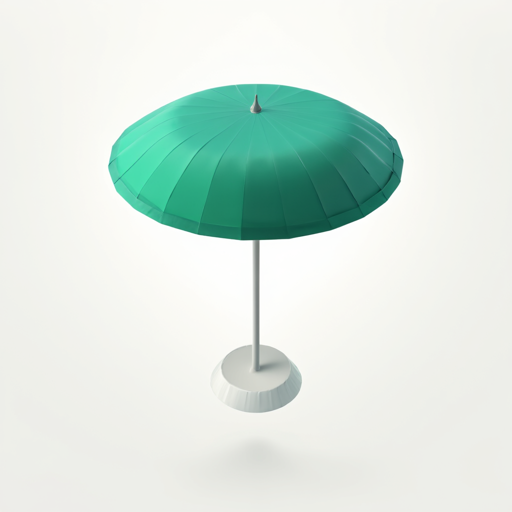} &
        \imageours{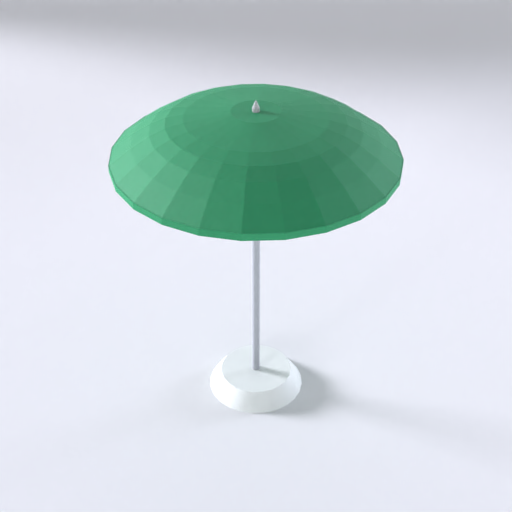} &
        \imageflux{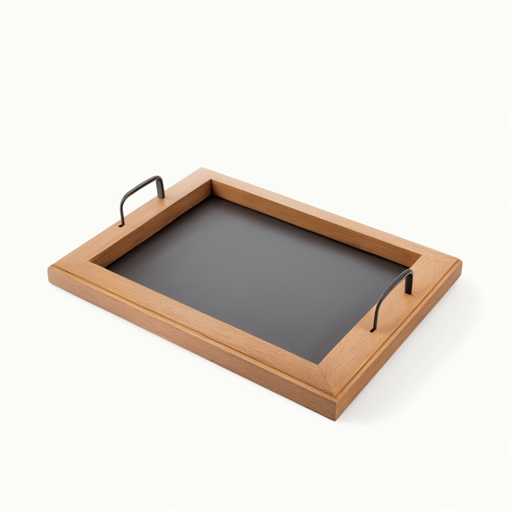} &
        \imageours{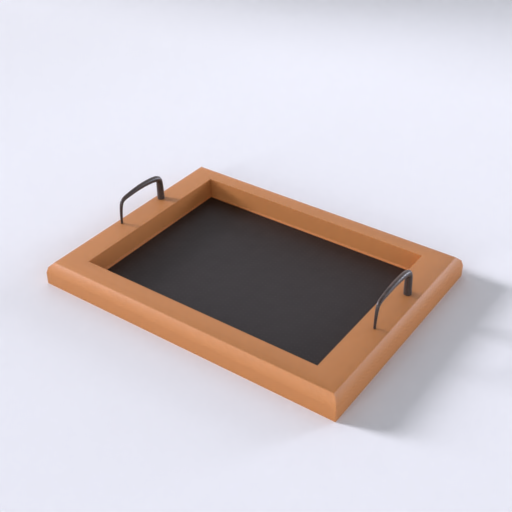} \\
        \imageflux{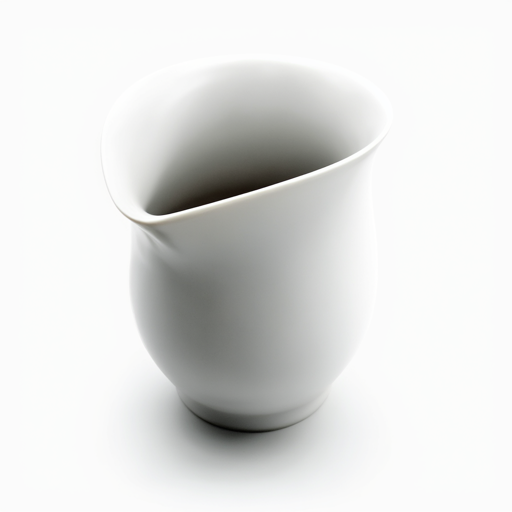} &
        \imageours{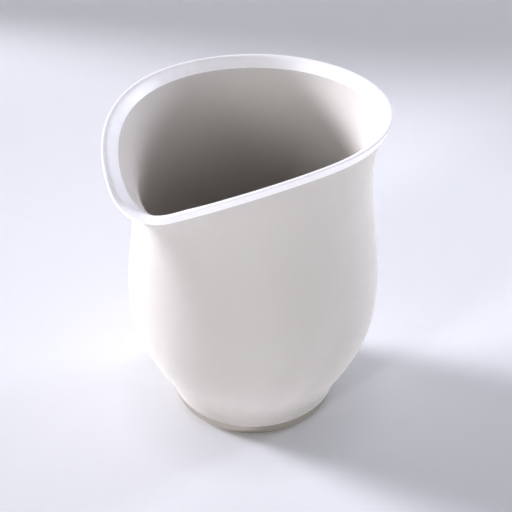} &
        \imageflux{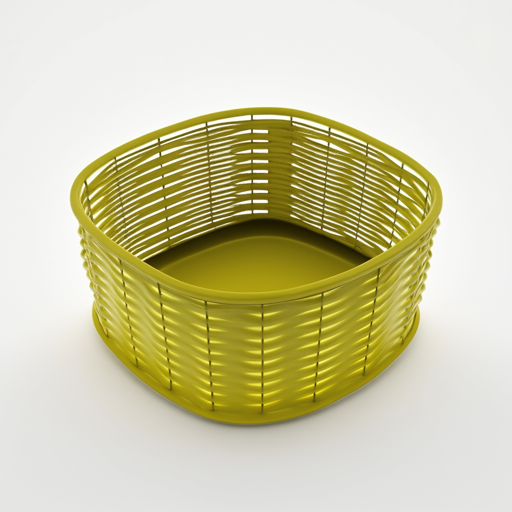} &
        \imageours{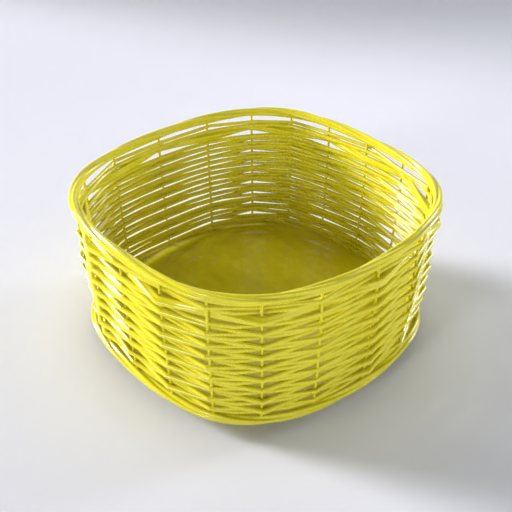} &
        \imageflux{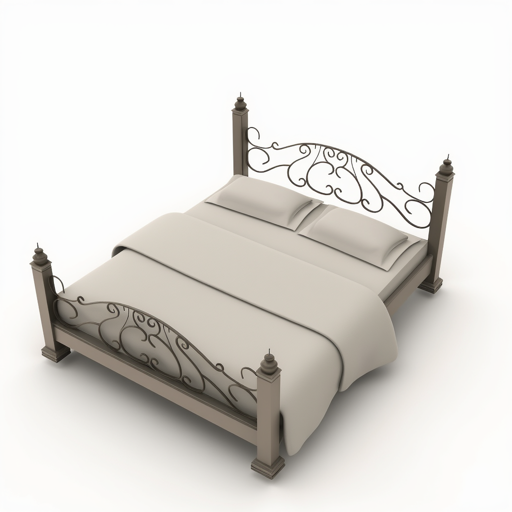} &
        \imageours{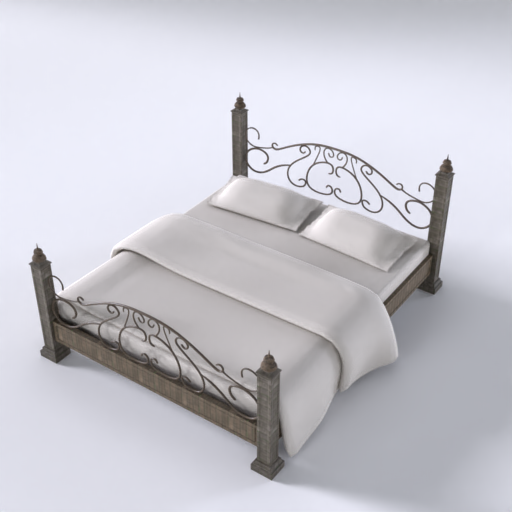}
	\end{tabular}
	\caption{Material matching results using MatCLIP on images of various shape categories from 3DCoMPaT++~\cite{li20223d_compat, slim_3dcompatplus_2023}, generated by Flux~\cite{flux2023}.}
	\label{fig:results_main_flux_mix}
\end{figure*}

\end{document}